\documentclass[pdflatex,sn-apa,iicol]{sn-jnl}% IJCV-recommended APA two-column format

\usepackage{graphicx}%
\usepackage{multirow}%
\usepackage{amsmath,amssymb,amsfonts}%
\usepackage{amsthm}%
\usepackage{mathrsfs}%
\usepackage{xcolor}%
\usepackage{colortbl}%
\usepackage{textcomp}%
\usepackage{manyfoot}%
\usepackage{booktabs}%
\usepackage{longtable}%
\usepackage{pifont}%
\usepackage{tikz}%
\usepackage{algorithm}%
\usepackage{algorithmicx}%
\usepackage{algpseudocode}%
\usepackage{listings}%
\usepackage{placeins}%

\definecolor{mmedarkgreen}{RGB}{0,128,0}
\definecolor{mmeamber}{RGB}{230,150,0}
\definecolor{mmegroupblue}{RGB}{232,240,249}
\definecolor{mmegroupgray}{RGB}{242,242,242}
\newcommand{\yesmark}{\textcolor{mmedarkgreen}{\ding{51}}}
\newcommand{\nomark}{\textcolor{red}{\ding{55}}}

\hypersetup{hypertexnames=false}

\theoremstyle{thmstyleone}%
\theoremstyle{thmstyletwo}%

\theoremstyle{thmstylethree}%

\begin{document}

\title[HUG-VIS]{HUG-VIS: A Multimodal Benchmark for Human-centered Understanding and Generation in Visual Intelligence}

\author[1]{\fnm{Fei} \sur{Ma}}

\author[1,2]{\fnm{Zebang} \sur{Cheng}}

\author[1]{\fnm{Minghui} \sur{Li}}

\author[1]{\fnm{Hongbo} \sur{Xu}}

\author[1,2]{\fnm{Yuyong} \sur{Tan}}

\author[3]{\fnm{Yihua} \sur{Shao}}

\author[4]{\fnm{Hanling} \sur{Wang}}

\author[1]{\fnm{Zhou} \sur{Liu}}

\author[5]{\fnm{Yuqing} \sur{Gao}}

\author[6]{\fnm{Dong} \sur{Wang}}

\author*[7]{\fnm{Long} \sur{Ma}}\email{longma@cuhk.edu.hk}

\author[2]{\fnm{Laizhong} \sur{Cui}}

\author[8]{\fnm{Nicu} \sur{Sebe}}

\author*[1,9]{\fnm{Qi} \sur{Tian}}\email{tian.qi1@huawei.com}

\affil[1]{\orgname{Guangdong Laboratory of Artificial Intelligence and Digital Economy (SZ)}, \orgaddress{\city{Shenzhen}, \country{China}}}

\affil[2]{\orgname{Shenzhen University}, \orgaddress{\city{Shenzhen}, \country{China}}}

\affil[3]{\orgname{Institute of Automation, Chinese Academy of Sciences}, \orgaddress{\city{Beijing}, \country{China}}}

\affil[4]{\orgname{Pengcheng Laboratory}, \orgaddress{\city{Shenzhen}, \country{China}}}

\affil[5]{\orgname{Tongji University}, \orgaddress{\city{Shanghai}, \country{China}}}

\affil[6]{\orgname{Tsinghua University}, \orgaddress{\city{Beijing}, \country{China}}}

\affil[7]{\orgname{The Chinese University of Hong Kong}, \orgaddress{\city{Hong Kong}, \country{China}}}

\affil[8]{\orgname{University of Trento}, \orgaddress{\city{Trento}, \country{Italy}}}

\affil[9]{\orgname{Huawei}, \orgaddress{\city{Shenzhen}, \country{China}}}

\abstract{
Visual intelligence aims to perceive, interpret, and synthesize the visual world, and it has become a central pursuit of modern computer vision. 
A particularly demanding branch of this field is human-centered visual intelligence, which studies people as expressive and socially situated subjects. Because human meaning is rarely carried by appearance alone, this branch increasingly analyzes people by coupling vision with audio and language, and its representative understanding and generation tasks span human emotion recognition, human video generation, human voice cloning, and human video matting. 
Despite rapid advancements in various tasks, existing data resources remain task-specific. They primarily provide the modalities and annotations needed to solve a single problem.
Therefore, the community still lacks a data foundation that coordinates and aligns different understanding and generation tasks. This hinders the full utilization of multimodal signals and impedes more comprehensive research into human-centered understanding and generation.
To address this gap, we present \textbf{HUG-VIS}, 
a unified benchmark for \textbf{H}uman-centered \textbf{U}nderstanding and \textbf{G}eneration in \textbf{VIS}ual intelligence.
We first build a controlled dataset of 8,400 seated half-body videos performed by 30 professional actors, each completing the same 280 emotion–action–prompt assignments
under a controlled Mandarin studio protocol, with synchronized video, audio, text, and alpha mattes. 
Building on this resource, we evaluate 
diverse
open- and closed-source models across the four tasks under a unified zero-shot protocol, pairing automatic metrics with criterion-specific mean opinion scores, and 
introduce multiple cross-task analyses.
The experiments show that \textbf{(i)} 
linguistic content dominates current emotion recognition, whereas purely visual affect recognition remains the weakest setting;
\textbf{(ii)} 
in both generation tasks (video generation and voice cloning), automatic metrics and human judgment agree in overall trend yet diverge in their top rankings, so the two dimensions must be reported jointly; 
\textbf{(iii)} boundary fidelity under motion is the principal remaining obstacle for human matting; 
and \textbf{(iv)} 
task difficulty varies with emotions, models, and metrics, with notable cross-task correlations.
The dataset and reported results are made available 
at~\url{https://github.com/GML-MMGroup/HUG-VIS}.
}


\maketitle

\section{Introduction}
\label{sec:introduction}

Visual intelligence has progressed remarkably from recognizing objects, scenes, and actions toward reasoning about and generating rich visual content, propelled by large-scale data and, more recently, multimodal foundation models \citep{ramesh2021zeroshot,li2024multimodalfoundation}. A particularly demanding frontier concerns \textit{people}: understanding what a person expresses and intends, and generating human appearance, motion, and voice in a controllable way. This \textbf{human-centered visual intelligence} is inherently multimodal, because facial expression, body motion, vocal prosody, and linguistic content provide complementary cues that are generally better interpreted jointly than in isolation \citep{banziger2012gemep,zadeh2018mosei}.

Along the path from understanding to generation, four capabilities have become representative. On the understanding side, multimodal emotion recognition infers affective and social states from visual, acoustic, and textual evidence \citep{ma2025generative}. On the generation side, human video generation animates a subject driven by audio signals or visual signals \citep{vougioukas2020speech,bounareli2024oneshot},
while voice cloning reproduces a target speaker's timbre and prosody from a limited enrollment sample \citep{ju2024naturalspeech}. Bridging the two ends, human video matting recovers temporally coherent alpha mattes that make captured foregrounds recomposable \citep{lin2021rvm,yang2025matanyone}.

Despite rapid progress, these capabilities are built and evaluated in isolation. 
Emotion recognition relies on corpora such as IEMOCAP and CMU-MOSEI \citep{busso2008iemocap,zadeh2018mosei}, video generation on MEAD and BEAT \citep{wang2020mead,liu2022beat}, voice cloning on VCTK \citep{yamagishi2019vctk}, and matting on VideoMatte240K \citep{lin2021rvm}. Each dataset provides only the modalities and annotations required by its target task, and differs in identity, behavior, and acquisition conditions. 

This fragmentation has two consequences. 
First, no corpus jointly offers emotion annotations, body-motion information, alpha supervision, and synchronized audio–video–text over the same subjects, which constrains progress along both directions \citep{liu2022beat,lin2021rvm}. 
Second, comparisons across disparate benchmarks conflate model capability with confounds such as sample population and evaluation protocol, so the connections among tasks cannot be reliably established \citep{tang2025humancentricfm}. 
These limitations point to the need for a human-centered data foundation that places understanding and generation on a common and aligned basis. 
Such a foundation can ensure that each task retains its own input, output, and metrics, while also supporting analysis across different tasks, thereby enabling more comprehensive research on human-centered understanding and generation.

To fill this gap in existing works, we introduce \textbf{HUG-VIS}, a condition-aligned benchmark for \textbf{H}uman-centered \textbf{U}nderstanding and \textbf{G}eneration in \textbf{VIS}ual intelligence, as shown in Fig.~\ref{fig:overview}.
It is the first benchmark to place multimodal emotion recognition, human video generation, voice cloning, and human video matting in a unified evaluation system.
The resource follows a complete actor-by-assignment design: 30 gender-balanced professional actors each perform an identical inventory of 280 emotion–action–prompt assignments, yielding 8,400 seated half-body clips captured under a controlled Mandarin studio protocol. 
The assignment space is organized as seven instructed emotions, including Happy, Angry, Sad, Afraid, Disgusted, Surprised, and Neutral, crossed with four emotion-consistent action templates per condition and ten scenario-based Mandarin utterances per action.
Every clip is packaged as a self-contained unit bundling synchronized RGB video, noise-suppressed audio, an assigned prompt and its verbatim transcript, and a sequence-level alpha matte.

Building on this resource, we conduct a comprehensive evaluation of diverse open- and closed-source systems across all four tasks under a unified zero-shot protocol.
The unified evaluation yields four findings.
First, current multimodal emotion recognition is dominated by linguistic content, whereas purely visual affect recognition remains by far the weakest setting. 
Second, in both generation tasks (video generation and voice cloning), the automatic metrics and human judgments agree in overall trend, but differ at the top.
Third, 
the principal remaining obstacle for video matting is maintaining boundary fidelity under motion, which is most evident along thin and rapidly deforming structures such as extended fingers and changing hand contours.
Fourth, the cross-task analysis reveals that task difficulty is not an intrinsic property but emerges jointly from the emotion, the model, and the metric. 
For example, recognition and matting errors show a negative correlation, the two voice-quality predictors are highly consistent, both audio-driven and vision-driven generation place equal emphasis on the quality of the generated video, and expansive emotions such as Happy, Angry, and Disgusted are the hardest for motion matting.
Collectively, by quantifying the difficulty of different understanding and generation tasks within a unified and condition-aligned coordinate system, these results provide a solid empirical basis for weighing the trade-offs among these tasks.

To sum up, this work makes three contributions:

\begin{figure*}[!t]
\centering
\includegraphics[width=\textwidth]{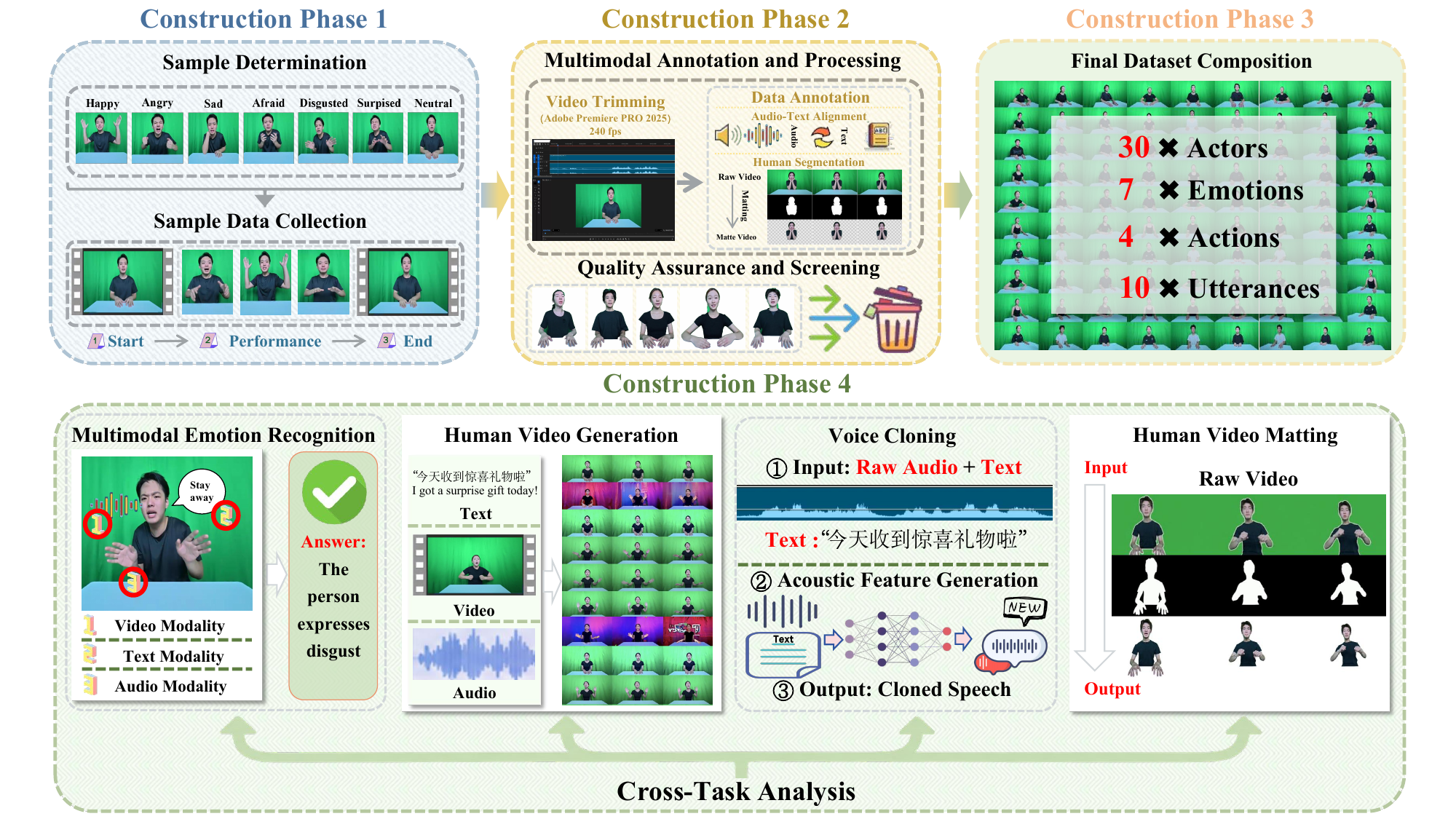}
\caption{
Overview of the HUG-VIS construction and evaluation pipeline. Phases 1–3 establish the shared grid and package temporally synchronized multimodal assets (RGB video, noise-suppressed audio, prompt and transcript, and alpha mattes), while 
Phase 4 conducts the evaluation of the four understanding and generation tasks together with the cross-task analysis.
}
\label{fig:overview}
\end{figure*}

\begin{itemize}
    \item 
    We construct HUG-VIS, the first condition-aligned human-centered dataset that  jointly offers emotion annotation, half-body capture, alpha supervision, and synchronized audio–video–text streams, comprising 8,400 clips from 30 actors spanning seven emotions, four actions per emotion, and ten utterances per action.
    \item 
    We evaluate the four understanding and generation tasks under a common zero-shot protocol and introduce a series of cross-task analyses, from which the above findings are derived.
    \item 
    We provide a detailed discussion of the insights and promising future research directions surrounding human-centered visual intelligence.
\end{itemize}

\section{Related Work}\label{sec:related_work}
Our benchmark is situated within the broader trajectory of human-centered visual intelligence and, more specifically, within the data resources that this field has relied upon.

\textbf{The scope of human-centered visual intelligence}. 
Human-centered visual intelligence studies people as structured, expressive, and socially situated subjects, connecting appearance, identity, motion, speech, and affect within a single object of study \citep{tang2025humancentricfm}. 
Early research concentrated on body-centered perception—re-identification, pose estimation, human parsing, attribute recognition, and crowd counting, exploiting the strong geometric priors that these tasks share, and increasingly unifying them under shared backbones and pretraining objectives \citep{ci2023unihcp}.
However, this perceptual foundation treats the person largely as a visual configuration to be localized and categorized, and stops short of the expressive and communicative dimensions that make human behavior meaningful.

Two developments have since broadened the field along orthogonal axes. 
Along the modeling axis, compact task-specific networks have given way to broadly pretrained, instruction-tuned models capable of recognizing temporally situated behavior and affect and of generating controllable human content \citep{li2024multimodalfoundation,stergiou2025abouttime,xing2026emollama,tang2025humancentricfm}.
Along the signal axis, the field has moved 
from vision alone toward the joint exploitation of vision, speech, language, and motion, since human meaning rarely resides in pixels in isolation and often depends on what is said, how it is said, and how it is enacted \citep{baltrusaitis2019multimodal,mago2026abstract,georgakis2018dynamic}.
Under these two shifts, understanding and generation cease to be separate agendas: a model that can infer a person's affective state from multimodal evidence draws on representations closely related to those needed to synthesize expressive appearance, motion, and voice. It is this convergence that motivates studying the two directions on a common footing rather than in isolation.

\textbf{Data resources and their fragmentation}. 
The empirical progress of the field has been tightly coupled to the resources that define its tasks, and this coupling is precisely where the field remains fragmented. 
On the understanding side, affective corpora such as IEMOCAP and CMU-MOSEI supply multimodal emotional data but offer no alpha supervision or controlled body-motion inventory \citep{busso2008iemocap,zadeh2018mosei}.
On the generation side, datasets such as MEAD and BEAT provide synchronized audio-visual material for animation, but they vary in their emotion annotations and in the extent of body coverage, and none is designed to support reuse across tasks \citep{wang2020mead,liu2022beat}.
Voice cloning has developed on multi-speaker corpora such as VCTK \citep{yamagishi2019vctk}, and video matting on large synthetic-composite collections such as VideoMatte240K \citep{lin2021rvm}, each optimized for its own reference signal.
More recent efforts to raise evaluation standards, including HumanVBench, sharpen the field's vocabulary but remain oriented toward a single capability or modality family \citep{zhou2026humanvbench}.

The impact is twofold. First, because each resource captures different identities under different conditions, comparisons drawn across them conflate genuine model capability with confounds of population, behavior, and acquisition. 
Second, and more fundamentally, 
no existing corpus supports the consistent evaluation of understanding and generation capabilities within a unified setting, and as a result, the relationships among these capabilities, such as whether their difficulty is consistent and whether their errors are correlated, remain entirely unestablished.
HUG-VIS is designed to remove exactly this obstacle: by having the same actors perform an identical emotion–action–prompt grid and by packaging synchronized video, audio, text, and alpha mattes for every performance, it places the four representative tasks on a shared and condition-aligned foundation, as detailed in the following Section.

\section{The Dataset}
\label{sec:benchmark}
Based on the aforementioned motivation, the organizational approach of HUG-VIS enables human performances to serve as evidence for both understanding and generation. 
Here, we introduce the dataset design principle, recording protocol, annotation specification, and final dataset composition.
\subsection{Design Principle: A Condition-Aligned Performance Grid}
\label{subsec:design}
HUG-VIS is built on a complete actor-by-assignment grid. Every actor performs an identical set of emotion–action–prompt assignments, so that visual behavior, speech, text, and foreground observations are aligned not only within a clip but also across actors and, later, across model outputs. 
This design deliberately trades environmental diversity for controlled correspondence: because the same assignment recurs for every actor, any difference we observe downstream can be attributed to the actor, the model, or the evaluation metric rather than to a mismatch in the underlying material. 
The evaluation unit is a single actor's performance of a single assignment. 
Each performance is linked to a synchronized set of modalities, comprising the RGB video and its accompanying audio, the assigned prompt together with its verbatim transcript, and the alpha matte. Retaining all 30 actors across all 280 assignments produces a complete grid of 8,400 performances.

\begin{table*}[!t]
\caption{
Emotion-conditioned performance design in HUG-VIS, showing the intended movement style, action templates, and an illustrative prompt for each emotion. 
The example prompts are English translations of the Mandarin prompts.
}
\label{tab:emotion_action_prompt_design}
\centering
\scriptsize
\setlength{\tabcolsep}{3pt}
\renewcommand{\arraystretch}{1.30}
\begin{tabular}{@{}>{\raggedright\arraybackslash}p{0.105\textwidth}p{0.20\textwidth}p{0.35\textwidth}p{0.27\textwidth}@{}}
\toprule
Condition & Movement style & Action templates & Illustrative prompt \\
\midrule
Happy & Light, relaxed, and expansive, with an open posture & Two-handed open-palmed welcome; one-handed open-palmed welcome; outward arm extension; two-handed thumbs-up gesture & ``I finally got a ticket to the concert! I'm thrilled!'' \\
\addlinespace[2pt]
Angry & Fast, forceful, and sustained, with a forward-oriented posture & Table strike; forceful arm sweep; sharp pointing gesture; clenched-fist raise and drop & ``Enough! I don't want to hear another excuse.'' \\
\addlinespace[2pt]
Sad & Slow or paused, with a contracted, downward-oriented posture & Head or neck support with one or both hands; hand on head; self-embrace with hands on upper arms & ``Everything I'd hoped for has fallen apart, and now I'm all alone.'' \\
\addlinespace[2pt]
Afraid & Contracted, guarded, and withdrawing, with slight trembling & One- or two-handed guarding gesture; recoil; hands clasped close to the body while leaning backward & ``Stay back! Please don't hurt me!'' \\
\addlinespace[2pt]
Disgusted & Restrained and outwardly rejecting, with a backward lean & Two-handed push-away; one-handed wave or push-away; dismissive wave; shooing gesture & ``Get this away from me! I can't stand looking at it.'' \\
\addlinespace[2pt]
Surprised & Rapid and moderately expansive, ending in an abrupt freeze & Sudden raising of both hands; hands brought behind the head; sudden torso lean; one-hand raise followed by a freeze & ``There was nothing behind the curtain. So why is that shadow moving?'' \\
\addlinespace[2pt]
Neutral & Gentle, even, and low-amplitude, with a stable, relaxed posture & Hands at rest with subtle finger or wrist movements; natural seated posture with minor head or hand adjustments & ``The meeting is at 3 p.m. Remember to bring the materials.'' \\
\botrule
\end{tabular}
\end{table*}

\subsection{Emotion, Action, and Utterance Design}
\label{subsec:emo_design}
The grid pairs six categories from the basic-emotion taxonomy \citep{ekman1992basic}, namely Happy, Angry, Sad, Afraid, Disgusted, and Surprised, together with Neutral in which actors are instructed to perform with minimal expressivity, giving seven conditions in total. 
For each emotion, 
the intended movement style, action templates, and an illustrative prompt are specified, as shown in Table~\ref{tab:emotion_action_prompt_design}. 
The action templates draw on established accounts of bodily emotion, movement dynamics, and action readiness \citep{demeijer1989movement,atkinson2004body,wallbott1998bodily,frijda1989action}, and the Mandarin prompts are drawn from acted-affect corpora following scenario-based emotion elicitation \citep{banziger2012gemep}. 
For each actor, the four action templates are crossed with ten textual prompts, yielding 40 assignments per emotion and thus 280 assignments across the seven conditions.

\begin{table*}[!t]
\caption{HUG-VIS acquisition setup and dataset composition.}
\label{tab:capture_spec}
\centering
\footnotesize
\setlength{\tabcolsep}{4pt}
\renewcommand{\arraystretch}{1.20}

\begin{minipage}[t]{0.49\textwidth}
\centering
\textbf{(a)~Acquisition setup}\par\smallskip
\renewcommand{\arraystretch}{1.25}
\begin{tabular}{@{}p{0.28\linewidth}p{0.66\linewidth}@{}}
\toprule
Item & Specification \\
\midrule
RGB camera &
DJI Action~5 Pro, fixed frontal mount \\
\addlinespace[1.5pt]

Background &
Uniform green screen for chroma-key matting \\
\addlinespace[1.5pt]

Lighting &
JHC-2000S LED, fixed color temperature and intensity \\
\addlinespace[1.5pt]

Framing &
Seated half-body, constant subject--camera distance \\
\addlinespace[1.5pt]

Resolution &
$1920\times1080$ \\
\addlinespace[1.5pt]

Frame rate &
Captured at 240~FPS, released at 30~FPS \\
\addlinespace[1.5pt]

Microphone &
DJI Mic Mini transmitter \\
\addlinespace[1.5pt]

Audio stream &
Synchronized with video and noise-suppressed \\
\bottomrule
\end{tabular}
\end{minipage}
\hfill
\begin{minipage}[t]{0.49\textwidth}
\centering
\textbf{(b)~Dataset composition}\par\smallskip
\renewcommand{\arraystretch}{1.25}
\begin{tabular}{@{}p{0.34\linewidth}p{0.60\linewidth}@{}}
\toprule
Property & Value \\
\midrule
Actors &
30, gender-balanced, aged approximately 20 years \\
\addlinespace[1.5pt]

Instructed conditions &
7 in total (6 basic emotions and Neutral) \\
\addlinespace[1.5pt]

Action templates &
4 per emotion \\
\addlinespace[1.5pt]

Canonical utterances &
10 per action, yielding 40 per emotion and 280 per actor \\
\addlinespace[1.5pt]

Performance clips &
8{,}400 in total \\
\addlinespace[1.5pt]

Text fields &
Assigned Mandarin prompt and verified transcript \\
\addlinespace[1.5pt]

Per-clip assets &
RGB video, audio, text, and alpha matte \\
\bottomrule
\end{tabular}
\end{minipage}

\end{table*}

\subsection{Acquisition: Actors, Studio, and Recording Protocol}
\label{subsec:acquisition}
Table~\ref{tab:capture_spec} (a) summarizes the recording configuration. Every performance is captured with a fixed frontal green-screen setup under controlled lighting and a seated half-body framing, which keeps the subject, viewpoint, and background constant across the entire grid.
Video is recorded at 1920 × 1080 and 240 FPS (released at 30 FPS), and accompanied by synchronized denoised audio.
30 professional actors completed the full grid and passed the retention criteria. 
This cohort is gender balanced, has a mean age of roughly 20 years, and contributed about four hours of recording per actor.

\begin{figure*}[!t]
\centering
\includegraphics[width=\textwidth]{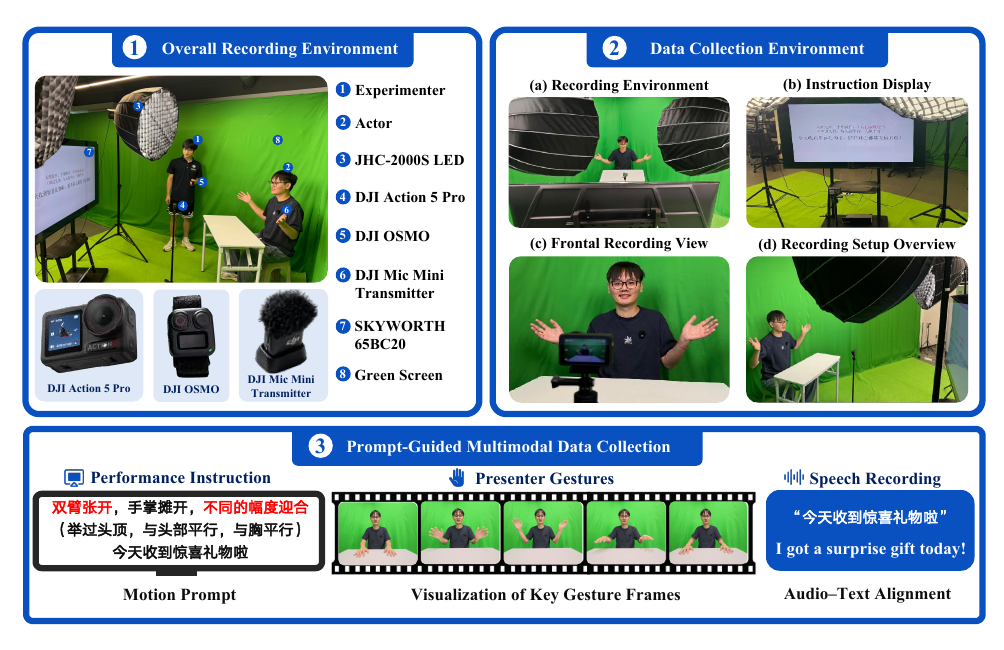}
\caption{
Acquisition setup of HUG-VIS: Region 1 shows the green-screen environment and equipment, Region 2 the front view configuration and instruction display, and Region 3 the prompt-guided recording with ordered key-gesture frames and synchronized speech. 
All clips are captured at 1920 $\times$ 1080 and 240 FPS (released at 30 FPS) with synchronized audio.
}
\label{fig:data_collection_environment}
\end{figure*}

Each performance follows a canonical protocol with three phases. The actor begins at rest, delivers the assigned prompt together with its action template, and then returns to the same resting pose. Because every clip shares these start, expression, and return phases, the protocol imposes consistent temporal boundaries that facilitate downstream processing and evaluation. Fig.~\ref{fig:data_collection_environment} illustrates the studio environment, the instruction display, and the synchronized capture of gesture and speech, and a side-view photograph documents the physical layout of the setup.

\subsection{Annotation and Quality Assurance}
Each evaluation unit comprises paired RGB video, audio, and text, together with the corresponding alpha mattes. 
The mattes are produced using the professional software Adobe Premiere Pro through a two-stage process: chroma-key initialization followed by sequence-level refinement. During refinement, particular attention is paid to hair, fingers, and clothing, as well as to regions affected by self-occlusion and motion blur.
All modalities are subsequently converted into a unified format, ensuring that the RGB, audio, text, and alpha channels of every unit follow consistent conventions.

\begin{figure*}[!t]
\centering
\includegraphics[width=\textwidth]{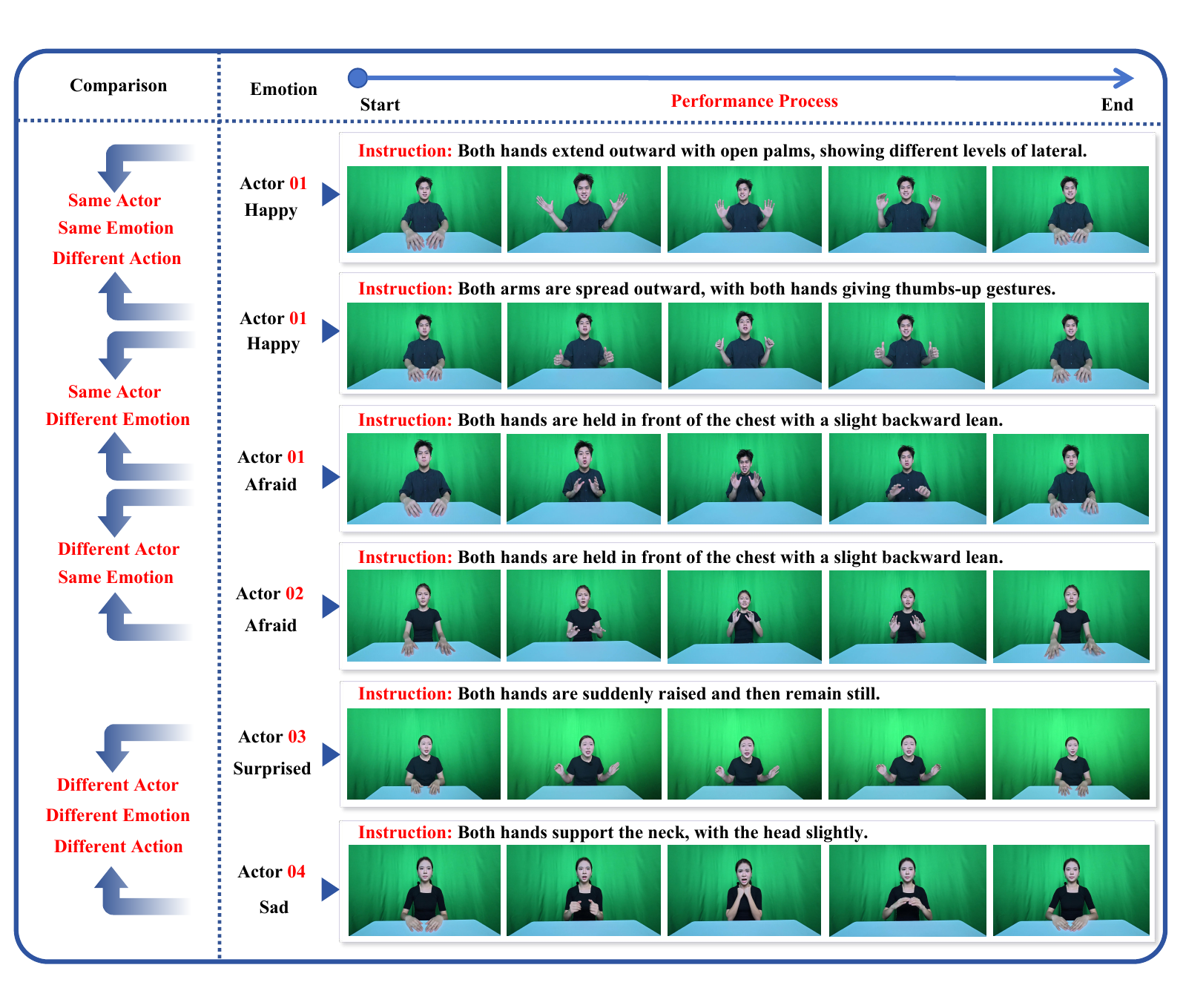}
\caption{
Samples illustrating the controlled correspondence of the actor-by-assignment grid, with each row showing ordered frames of the rest–expression–rest protocol. The indicated row pairs vary a factor at a time (actor, emotion, action, or all three), so that downstream differences are attributable to the actor, model, or criterion rather than mismatched material.
}
\label{fig:raw_data_examples}
\end{figure*}

To ensure the accuracy of the retained data, eight professional volunteers conduct a quality assessment. Each video clip is reviewed to verify strict alignment among the RGB, audio, text, and alpha mattes. In addition, the volunteers check audio-visual synchronization, the accuracy of the textual descriptions, and the presence of any segmentation errors caused by motion blur or self-occlusion. Any clip that fails to meet these criteria is either discarded or returned for reprocessing, thereby ensuring the overall quality and consistency of the final dataset.
Fig.~\ref{fig:raw_data_examples} shows example videos covering different actors, emotions, and actions.

\subsection{Dataset Composition and Comparison}
As illustrated in Table~\ref{tab:capture_spec} (b), the released dataset comprises 8,400 seated half-body clips that are evenly distributed across the seven emotions, with 1,200 clips per condition. Every clip is packaged as a self-contained evaluation unit that bundles the RGB video, the synchronized audio, the assigned text together with its transcript, and alpha mattes. 
Because these assets are temporally synchronized and share consistent conventions, a clip can be used directly for understanding and generation tasks.

\shortcites{zhang2021hdtf}

\begin{table*}[!t]
\caption{
Comparison of HUG-VIS with representative datasets along five properties: emotion annotation, half-body
capture, alpha supervision, audio–video–text (A-V-T) alignment, and condition-aligned cross-task analysis.
\yesmark\ denotes that the released resource satisfies the stated criterion, while \nomark\ denotes that it does not.
}
\label{tab:dataset_comparison}
\centering
\scriptsize
\setlength{\tabcolsep}{2.4pt}
\renewcommand{\arraystretch}{1.12}
\begin{tabular}{@{}>{\raggedright\arraybackslash}m{0.33\textwidth}>{\centering\arraybackslash}m{0.10\textwidth}>{\centering\arraybackslash}m{0.10\textwidth}>{\centering\arraybackslash}m{0.10\textwidth}>{\centering\arraybackslash}m{0.12\textwidth}>{\centering\arraybackslash}m{0.18\textwidth}@{}}
\toprule
\multicolumn{1}{c}{Resource} & \shortstack{Emotion\\annotation} & \shortstack{Half-body\\capture} & \shortstack{Alpha\\supervision} & \shortstack{A/V/T\\alignment} & \shortstack{Condition-aligned\\cross-task analysis} \\
\midrule
MEAD~\citep{wang2020mead} & \yesmark & \nomark & \nomark & \yesmark & \nomark \\
HDTF~\citep{zhang2021hdtf} & \nomark & \nomark & \nomark & \nomark & \nomark \\
BEAT~\citep{liu2022beat} & \yesmark & \yesmark & \nomark & \nomark & \nomark \\
IEMOCAP~\citep{busso2008iemocap} & \yesmark & \yesmark & \nomark & \yesmark & \nomark \\
VCTK~\citep{yamagishi2019vctk} & \nomark & \nomark & \nomark & \nomark & \nomark \\
VideoMatte240K~\citep{lin2021bgmv2} & \nomark & \yesmark & \yesmark & \nomark & \nomark \\
\addlinespace[1pt]
\midrule
\textbf{HUG-VIS} & \yesmark & \yesmark & \yesmark & \yesmark & \yesmark \\
\bottomrule
\end{tabular}
\end{table*}

Table~\ref{tab:dataset_comparison} compares HUG-VIS with representative
existing resources
\citep{wang2020mead,zhang2021hdtf,liu2022beat,busso2008iemocap,
yamagishi2019vctk,lin2021bgmv2}
under the five binary criteria defined in the caption.
These criteria characterize released assets and analysis protocols rather
than the overall quality or scope of each resource.
MEAD and IEMOCAP provide emotion-labeled A/V/T data; HDTF provides
audio-visual talking-face clips but lacks paired transcripts; BEAT provides
emotion-labeled body-motion data but not standardized upper-body RGB video;
VCTK provides speech and utterance transcripts but lacks paired video; and
VideoMatte240K provides human foreground sequences with per-frame alpha
mattes but uses heterogeneous footage rather than a standardized half-body
capture protocol.
Among the resources compared, and under these criteria, HUG-VIS is the only
resource that satisfies all five.
Its complete actor-by-assignment grid and shared actor, source, and condition
identifiers allow task-specific results to be analyzed on matched identities
and instructed conditions.

\section{Benchmark Protocols and Metrics}
\label{sec-protocol}
\subsection{Tasks and Comparison Populations}
\label{sec-Tasks}
HUG-VIS defines four tasks, each with its own input, target, reference, comparison population, and evaluation metrics. 
All tasks follow a zero-shot evaluation protocol, so no benchmark samples are used for training, fine-tuning, calibration, or model selection.
% All tasks use zero sample evaluation, so the benchmark sample will not enter training, fine-tuning, calibration, or model selection.
\textbf{Multimodal emotion recognition} asks systems to predict the assigned emotion label from image frames, video, audio, text, or different combinations thereof, and results are reported separately. 
\textbf{Human video generation} is organized by driving signal into audio-driven and vision-driven regimes, and both are included in our evaluation. 
Audio-driven generation synthesizes performances from the recorded audio, whereas vision-driven generation is conditioned on visual references, with each system evaluated according to its reported output scope. 
\textbf{Voice cloning} reproduces the vocal identity of a target speaker in synthesized speech, and each output is tied to its actor 
so that it can be compared directly with the recorded reference audio of the same performance.
\textbf{Human video matting} recovers the  alpha matte from the RGB frames and is evaluated against the reference alpha, which measures both spatial accuracy and temporal stability.

\subsection{Objective and Subjective Evaluation Metrics}
\label{subsec:task_evaluation}

Each task is evaluated with metrics matched to its output representation and reference signal.
Emotion recognition is measured by seven-class classification accuracy. 
Audio-driven generation reports identity similarity via ArcFace cosine similarity \citep{deng2019arcface} (CSIM) and lip-synchronization quality via SyncNet-derived confidence and distance measures (denoted Sync-C and Sync-D) \citep{chung2016out,prajwal2020wav2lip,ji2025sonic,tu2025stableavatar}, whereas vision-driven generation reports LPIPS \citep{zhang2018lpips}, CSIM, PSNR, SSIM \citep{wang2004ssim}, and FID \citep{heusel2017fid}.
Voice cloning is evaluated using UTMOS \citep{saeki2022utmos} and DNSMOS \citep{reddy2021dnsmos} as reference-free speech-quality predictors, together with Resemblyzer-based speaker similarity \citep{wan2018generalized,resembleai2026resemblyzer} as a measure of identity preservation.
Matting reports spatial MAD, MSE, gradient, and connectivity errors against the reference alpha, together with the temporal dtSSD \citep{erofeev2015video}.

Beyond these objective metrics, we conduct subjective studies that report criterion-specific mean opinion score (MOS) on a 1–5 scale. 
For video generation, human raters assess identity preservation, emotion naturalness, and either lip synchronization for the audio-driven setting or motion naturalness for the vision-driven setting. 
For voice cloning, raters listen to each generated utterance alongside its reference audio and score their similarity, as well as the emotion naturalness of the generated speech. Ratings are averaged within each clip to yield the MOS.

\subsection{Cross-Task Analysis Setup}
\label{subsec:cross_task_analysis_setup}
The four benchmark tasks share the same actor and source, 
which allows us to study how difficulty varies across tasks rather than within a single one. 
However, because the metrics of Section~\ref{subsec:task_evaluation} differ in unit and optimization direction, their raw values cannot be compared directly. Therefore, 
we propose four cross-task analysis metrics, whose corresponding results are presented in Section~\ref{sec:cross_task_analysis}.

\textbf{Emotion difficulty across metrics}. 
This analysis compares the results of different emotions within the same evaluation metric to demonstrate the role of emotions in the final outcome.
Let \(\mathcal{E}\) represent the set of seven emotions. For each emotion \(e\in\mathcal{E}\), task metric \(k\), and model \(m\in\mathcal{M}_k\) (the set of models evaluated under metric \(k\)), 
let \(g_{m,e,k}\) denote the sample-averaged value of model  \(m\) under condition \(e\) with respect to the metric \(k\), oriented so that a larger value indicates greater difficulty. Then, we average the models:
\begin{equation}
g_{e,k}
=
\frac{1}{|\mathcal{M}_k|}
\sum_{m\in\mathcal{M}_k} g_{m,e,k}
\end{equation}

Then, each \(g_{e,k}\) is min-max normalized over \(\mathcal{E}\),
\begin{equation}
\begin{aligned}
D_{e,k} = \frac{g_{e,k} - g_k^{\min}}{g_k^{\max} - g_k^{\min}} \in [0,1]
\end{aligned}
\label{eq:cross_task_difficulty}
\end{equation}
so that 0 and 1 mark the easiest and hardest condition under metric \(k\), respectively. 
The results are visualized in Fig.~\ref{fig:emotion_metric_difficulty}.

\textbf{Agreement between difficulty profiles}. 
This analysis computes the metric agreement of the emotion difficulty described above.
For metric \(k\), the normalized values form a profile \(D_k = (D_{e,k})_{e \in \mathcal{E}}\), and the agreement between metrics \(k\) and \(\ell\) can be computed as Spearman rank correlation:
\begin{equation}
\rho_{k,\ell} = \operatorname{Spearman}(D_k, D_\ell)
\label{eq:cross_AGREEMENT}
\end{equation}
where positive, negative, and near-zero values indicate positive correlation, negative correlation, and no correlation, respectively.
The agreement matrix is reported in Fig.~\ref{fig:metric_profile_agreement}.

\textbf{Source-level difficulty across generation regimes}. 
This analysis aims to explore whether audio-driven video generation (AD) and vision-driven video generation (VD) impose similar relative difficulty on the same source clips.
For model \(m\), clip \(s\), and metric \(q\), 
each metric value \(z_{m,s}^{(q)}\) 
is first min–max normalized across all selected clips, yielding a higher-is-harder score.
Then, the mean of these normalized values across different metrics gives the difficulty \(z_{m,s}\) that model \(m\) assigns to clip \(s\).
By averaging different models within each regime of AD and VD, we can obtain
\begin{equation}
Z_s^p = \frac{1}{|\mathcal{M}_p|} \sum_{m \in \mathcal{M}_p} z_{m,s}, \qquad p \in \{\mathrm{AD}, \mathrm{VD}\}
\label{eq:cross_soure_level}
\end{equation}

Then, we plot \(Z_s^{\mathrm{AD}}\) against \(Z_s^{\mathrm{VD}}\) for each sample in a single coordinate system and fit the resulting distribution, as shown in Fig.~\ref{fig:head_synthesis_agreement}.

\textbf{Motion difficulty in the matting task and its relationship with emotions}.
This analysis aims to explore the motion challenges in video matting tasks and analyze the motion difficulty of different emotions by fitting the results to a matting evaluation metric.
Specifically, inspired by previous matting works
\citep{erofeev2015video,johnson2016sparse,perazzi2016benchmark}, we define a motion difficulty that combines frame-to-frame alpha variation with normalized foreground displacement. In the formalization, 
we first let \(\alpha_{s,t}(x,y) \in [0,1]\) be the alpha value at spatial coordinate \((x,y)\), \(1 \le x \le W,\ 1 \le y \le H\), in the \(t\)-th of the \(T_s\) sampled frames of clip \(s\), each of size \(H\times W\).
Then, we summarize local variation by the mean absolute inter-frame alpha change \(U_s\) and global displacement by the mean inter-frame shift \(V_s\) of the normalized foreground centroid,
\begin{equation}
\resizebox{\columnwidth}{!}{$
U_s = \frac{1}{(T_s-1)HW} \sum_{t=1}^{T_s-1} \sum_{x,y} |\alpha_{s,t+1}(x,y) - \alpha_{s,t}(x,y)|
$}
\label{eq:motion1}
\end{equation}
\begin{equation}
\qquad V_s = \frac{1}{T_s-1} \sum_{t=1}^{T_s-1} \| c_{s,t+1} - c_{s,t} \|_2
\label{eq:motion2}
\end{equation}
where the centroid \(c_{s,t}\) is computed from the thresholded foreground mask. 
Because \(U_s\) and \(V_s\) differ in scale, each is converted to a percentile rank and their mean defines the motion difficulty \(D_{\mathrm{motion}}(s)\), a quantity that depends on the reference alpha values. 
Next, we average the matting metric \(\mathrm{dtSSD}\) over the different models for each clip \(s\), yielding \(\mathrm{dtSSD}_s\). We fit the relationship between \(D_{\mathrm{motion}}(s)\) and \(\mathrm{dtSSD}_s\) in a coordinate system, annotating the samples according to their instructed emotion, as shown in Fig.~\ref{fig:gt_difficulty_mechanisms}.

These four cross-task analyses characterize difficulty in terms of instructed emotion, source clip, and video motion. 
In Section~\ref{sec:cross_task_analysis}, they offer a principled basis for diagnosing where current models break down and for comparing methods along axes that single-task metrics fail to capture.

\section{Experimental Results}\label{sec:experiments}
Following the zero-shot protocol of Section~\ref{sec-protocol}, 
we report evaluation results for each individual task and across tasks.

\subsection{Multimodal Emotion Recognition}
\label{sec:emotion_recognition_results}

\begin{table*}[!t]
\caption{
Seven-class emotion recognition accuracy. Bold and underlined values indicate the best and second-best results.
}\label{tab:emotion_classification}
\centering
\scriptsize
\setlength{\tabcolsep}{2pt}
\renewcommand{\arraystretch}{0.82}
\begin{tabular*}{0.95\textwidth}{@{\extracolsep\fill}p{0.54\textwidth}p{0.28\textwidth}c@{}}
\toprule
Model & Input & Acc. (\%)$\uparrow$ \\
\midrule
\rowcolor{mmegroupblue}
\multicolumn{3}{@{}l}{\textbf{Image frame}} \\
\midrule
Former-DFER~\citep{zhao2021formerdfer} & Image frame & 11.42 \\
Emotion-FAN~\citep{meng2019fan} & Image frame & 13.90 \\
EmotiEffLib~\citep{emotiefflib2025,savchenko2023adaptive} & Image frame & \underline{33.37} \\
MMA-DFER~\citep{chumachenko2024mmadfer} & Image frame & \textbf{37.05} \\
\addlinespace[1pt]
\rowcolor{mmegroupgray}
\multicolumn{3}{@{}l}{\textbf{Video}} \\
\midrule
Qwen2.5-Omni-7B~\citep{qwen2025qwen25omni} & Video & 27.77 \\
HumanOmni-7B~\citep{zhao2025humanomni} & Video & 41.90 \\
VideoLLaMA3-7B~\citep{zhang2025videollama3} & Video & 43.12 \\
MiniCPM-V 4.5~\citep{yu2025minicpmv45} & Video & 45.88 \\
Molmo2-8B~\citep{clark2026molmo2} & Video & 46.34 \\
InternVideo2.5-Chat-8B~\citep{wang2025internvideo25} & Video & \underline{46.42} \\
MiniCPM-o 4.5~\citep{cui2026minicpmo45} & Video & \textbf{48.52} \\
\addlinespace[1pt]
\rowcolor{mmegroupblue}
\multicolumn{3}{@{}l}{\textbf{Audio}} \\
\midrule
Kimi-Audio-7B-Instruct~\citep{kimiteam2025kimiaudio} & Audio & 64.64 \\
Qwen2-Audio-7B-Instruct~\citep{chu2024qwen2audio} & Audio & 69.85 \\
EmotionThinker~\citep{wang2026emotionthinker} & Audio & \underline{72.98} \\
Audio-Reasoner-7B~\citep{xie2025audioreasoner} & Audio & \textbf{74.38} \\
\addlinespace[1pt]
\rowcolor{mmegroupgray}
\multicolumn{3}{@{}l}{\textbf{Text}} \\
\midrule
EmoBERTa~\citep{kim2021emoberta} & Text & 19.35 \\
StructBERT~\citep{wang2020structbert} & Text & 38.28 \\
Qwen2.5-7B-Instruct~\citep{qwen2024qwen25} & Text & 78.32 \\
Qwen3-32B~\citep{qwen2025qwen3} & Text & \underline{82.45} \\
DeepSeek-V3.2~\citep{deepseekai2025deepseekv32} & Text & \textbf{82.93} \\
\addlinespace[1pt]
\rowcolor{mmegroupblue}
\multicolumn{3}{@{}l}{\textbf{Video + Audio}} \\
\midrule
MiniCPM-o 4.5~\citep{cui2026minicpmo45} & Video + Audio & 47.07 \\
HumanOmni-7B~\citep{zhao2025humanomni} & Video + Audio & \underline{70.16} \\
Qwen2.5-Omni-7B~\citep{qwen2025qwen25omni} & Video + Audio & \textbf{73.70} \\
\addlinespace[1pt]
\rowcolor{mmegroupgray}
\multicolumn{3}{@{}l}{\textbf{Video + Text}} \\
\midrule
VideoLLaMA3-7B~\citep{zhang2025videollama3} & Video + Text & 54.34 \\
HumanOmni-7B~\citep{zhao2025humanomni} & Video + Text & 72.41 \\
Molmo2-8B~\citep{clark2026molmo2} & Video + Text & 75.23 \\
MiniCPM-V 4.5~\citep{yu2025minicpmv45} & Video + Text & 76.52 \\
MiniCPM-o 4.5~\citep{cui2026minicpmo45} & Video + Text & 77.12 \\
InternVideo2.5-Chat-8B~\citep{wang2025internvideo25} & Video + Text & \underline{78.77} \\
Qwen2.5-Omni-7B~\citep{qwen2025qwen25omni} & Video + Text & \textbf{83.74} \\
\addlinespace[1pt]
\rowcolor{mmegroupblue}
\multicolumn{3}{@{}l}{\textbf{Video + Audio + Text}} \\
\midrule
Qwen2.5-Omni-3B~\citep{qwen2025qwen25omni} & Video + Audio + Text & 58.41 \\
HumanOmni-7B~\citep{zhao2025humanomni} & Video + Audio + Text & 73.15 \\
MiniCPM-o 4.5~\citep{cui2026minicpmo45} & Video + Audio + Text & \underline{77.30} \\
Qwen2.5-Omni-7B~\citep{qwen2025qwen25omni} & Video + Audio + Text & \textbf{83.79} \\
\botrule
\end{tabular*}
\end{table*}

Table~\ref{tab:emotion_classification} reports seven-class recognition accuracy across the image, video, audio, text, and multimodal input settings, from which two findings emerge. First, under the zero-shot protocol, the compact task-specific baselines transfer poorly to this regime, while broadly pretrained and instruction-tuned models prove substantially more capable. 
For text-only input, accuracy ranges from 19.35\% with EmoBERTa to 82.93\% with DeepSeek-V3.2, whereas for audio-only input, it ranges from 64.64\% with Kimi-Audio-7B-Instruct to 74.38\% with Audio-Reasoner-7B.
Second, and most consequential for a human-centered benchmark, purely visual recognition remains the weakest setting by a wide margin.
The strongest image-frame and video systems reach only 37.05\% and 48.52\%, respectively, and although the temporal cues available in video yield an 11.47\% gain over single frames,
the best video result still trails the best audio result by 25.86\% and the best text result by 34.41\%. 
Consequently, the principal bottleneck on our dataset lies in inferring the emotion information from visual evidence alone, which demands sensitivity to subtle facial cues and their temporal dynamics.

\begin{table*}[!t]
\caption{Objective results of the task of audio-driven video generation. Bold and underlined values indicate the best and second-best results.}
\label{tab:audio_driven_head_synthesis}
\centering
\scriptsize
\setlength{\tabcolsep}{5pt}
\renewcommand{\arraystretch}{1.02}
\begin{tabular*}{\textwidth}{@{\extracolsep\fill}>{\raggedright\arraybackslash}p{0.43\textwidth}*{3}{>{\centering\arraybackslash}p{0.13\textwidth}}@{}}
\toprule
Model & CSIM$\uparrow$ & Sync-C$\uparrow$ & Sync-D$\downarrow$ \\
\midrule
EDTalk~\citep{tan2024edtalk} & 0.710 & 3.67 & 8.90 \\
V-Express~\citep{wang2024vexpress} & 0.786 & 4.95 & \underline{8.10} \\
EchoMimicV3~\citep{meng2026echomimicv3} & 0.771 & 3.15 & 10.30 \\
AniTalker~\citep{liu2024anitalker} & 0.744 & 3.51 & 9.93 \\
Hallo2~\citep{cui2024hallo2} & \underline{0.868} & 4.65 & 8.43 \\
Ditto~\citep{li2025ditto} & \textbf{0.904} & 3.47 & 9.31 \\
Sonic~\citep{ji2025sonic} & 0.736 & \underline{5.42} & \textbf{7.73} \\
LatentSync~\citep{li2024latentsync} & 0.798 & \textbf{5.43} & \textbf{7.73} \\
\botrule
\end{tabular*}
\end{table*}

\begin{table*}[!t]
\caption{
Subjective results of the task of audio-driven video generation. Bold and underlined values indicate the best and second-best results.
}
\label{tab:audio_driven_human_mos}
\centering
\scriptsize
\setlength{\tabcolsep}{3pt}
\renewcommand{\arraystretch}{0.96}
\begin{tabular*}{\textwidth}{@{\extracolsep\fill}>{\raggedright\arraybackslash}p{0.40\textwidth}>{\centering\arraybackslash}p{0.14\textwidth}*{2}{>{\centering\arraybackslash}p{0.20\textwidth}}@{}}
\toprule
Model & \makebox[\linewidth][c]{ID Similarity $\uparrow$} & \makebox[\linewidth][c]{Emotion Naturalness $\uparrow$} & \makebox[\linewidth][c]{Lip Synchronization $\uparrow$} \\
\midrule
EDTalk~\citep{tan2024edtalk} & 1.22 & 1.24 & 1.33 \\
V-Express~\citep{wang2024vexpress} & 1.36 & 1.33 & 3.40 \\
EchoMimicV3~\citep{meng2026echomimicv3} & 2.93 & 2.07 & 2.24 \\
AniTalker~\citep{liu2024anitalker} & 3.74 & 2.89 & 2.78 \\
Hallo2~\citep{cui2024hallo2} & 3.69 & 2.91 & 2.96 \\
Ditto~\citep{li2025ditto} & 4.13 & 3.02 & 2.94 \\
Sonic~\citep{ji2025sonic} & \textbf{4.44} & \textbf{4.16} & \underline{4.43} \\
LatentSync~\citep{li2024latentsync} & \underline{4.29} & \underline{4.13} & \textbf{4.47} \\
\botrule
\end{tabular*}
\end{table*}

The multimodal settings further illuminate how current models combine evidence across modalities. For Qwen2.5-Omni-7B, augmenting video with text lifts accuracy from 27.77\% to 83.74\%, yet the subsequent addition of audio to the video–text input contributes a mere 0.05 point. 
HumanOmni-7B and MiniCPM-o 4.5 follow the same trajectory, gaining 30.51\% and 28.60\% from text but only 0.74\% and 0.18\% from the further inclusion of audio. 
Particularly, for MiniCPM-o 4.5, accuracy even declines from 48.52\% with video alone to 47.07\% once audio is added. 
Collectively, these results establish linguistic content as the dominant source of evidence for current multimodal models on this dataset and reveal that the remaining modalities are not yet integrated into complementary gains. 
Continued progress on human-centered understanding will thus demand both more discriminative and fine-grained visual representations of affect and cross-modal fusion mechanisms capable of extracting genuinely complementary information from the audio and visual channels.

\begin{figure*}[!t]
\centering
\includegraphics[width=0.88\textwidth]{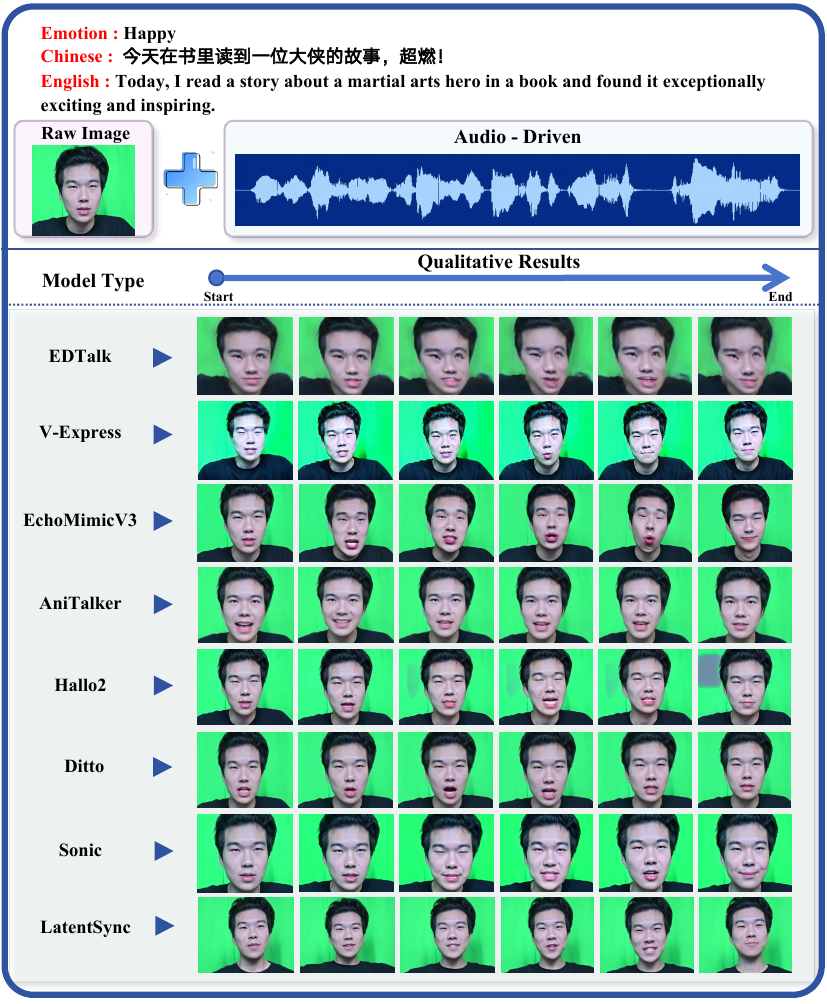}
\caption{
Qualitative comparison of audio-driven video generation for one Happy utterance, sampled at six ordered time points. The top panel shows the source portrait, transcript, and driving audio waveform, and each row shows the temporally aligned outputs of one model.
}
\label{fig:audio_driven_expression_sequence}
\end{figure*}

\subsection{Human Video Generation}
\label{sec:digital_human_synthesis_results}
We evaluate human video generation under two driving regimes. 
Audio-driven generation animates a portrait from recorded audio and is assessed for identity preservation and audio–visual synchronization, whereas vision-driven generation transfers motion from a driving video and is assessed for appearance reconstruction, identity consistency, and motion naturalness. 
In addition, following the vision-driven generation results reported in the original papers of each method, we organize the open-source models into two paradigms according to the spatial extent they are designed to animate: Head Animation systems reconstruct a driven portrait and concentrate on facial geometry and identity, whereas Body Animation systems render the body and should additionally coordinate torso and hand motion.

\textbf{Audio-driven generation}. The automatic results in Table~\ref{tab:audio_driven_head_synthesis} show that the models distribute their strengths across identity and synchronization rather than excelling on both at once. 
Ditto attains the highest identity similarity, with a CSIM of 0.904, while LatentSync records the best Sync-C at 5.43 and shares the best Sync-D of 7.73 with Sonic. 
The subjective study in Table~\ref{tab:audio_driven_human_mos} is broadly consistent with these measurements while refining their ordering. LatentSync obtains the highest lip-synchronization score of 4.47, in line with its leading Sync-C, so the two evaluations agree on synchronization quality. 
On the identity dimension, the rankings shift. Sonic leads both identity similarity and emotion naturalness, at 4.44 and 4.16, even though its CSIM reaches 0.736, whereas Ditto, the CSIM leader, falls to third in perceived identity. 
This pattern suggests that CSIM captures one aspect of identity, while human raters may also consider other factors such as stability and expressiveness.
The qualitative comparison in Fig.~\ref{fig:audio_driven_expression_sequence} makes this concrete. EDTalk exhibits severe facial deformation and unstable geometry, Hallo2 retains the source identity but introduces visible background and texture artifacts, and Sonic and LatentSync produce the most stable identities and expression dynamics.

\begin{table*}[!t]
\caption{
Objective results of the task of vision-driven video generation. Bold and underlined values indicate the best and second-best results.
}
\label{tab:vision_driven_synthesis}
\centering
\scriptsize
\setlength{\tabcolsep}{3pt}
\renewcommand{\arraystretch}{0.95}
\begin{tabular*}{\textwidth}{@{\extracolsep\fill}>{\raggedright\arraybackslash}p{0.40\textwidth}*{5}{>{\centering\arraybackslash}p{0.10\textwidth}}@{}}
\toprule
Model & LPIPS$\downarrow$ & CSIM$\uparrow$ & PSNR$\uparrow$ & SSIM$\uparrow$ & FID$\downarrow$ \\
\midrule
\rowcolor{mmegroupblue}
\multicolumn{6}{@{}l}{\textbf{Open-source}} \\
\midrule
\multicolumn{6}{@{}l}{\textit{Head Animation}} \\
HunyuanPortrait~\citep{xu2025hunyuanportrait} & 0.568 & \underline{0.868} & 9.11 & 0.375 & 175.71 \\
X-NeMo~\citep{zhao2025xnemo} & \textbf{0.520} & 0.678 & \textbf{10.15} & 0.368 & \textbf{150.57} \\
\shortcites{li2026personalive}% Keep the author list abbreviated in both adjacent tables.
PersonaLive!~\citep{li2026personalive} & \underline{0.565} & 0.834 & \underline{9.43} & \underline{0.382} & \underline{164.94} \\
AniPortrait~\citep{wei2024aniportrait} & \underline{0.565} & \textbf{0.884} & \underline{9.43} & \textbf{0.383} & 167.49 \\
\addlinespace[1pt]
\multicolumn{6}{@{}l}{\textit{Body Animation}} \\
MimicMotion~\citep{zhang2025mimicmotion} & 0.606 & 0.627 & 9.24 & 0.557 & 59.56 \\
StableAnimator~\citep{tu2025stableanimator} & 0.280 & 0.695 & 14.52 & 0.601 & 55.99 \\
Animate-X~\citep{tan2025animatex} & \textbf{0.139} & \underline{0.733} & \textbf{18.81} & \textbf{0.755} & \underline{31.46} \\
Wan2.2~\citep{wanvideo2025wan22} & \underline{0.195} & \textbf{0.783} & \underline{16.08} & \underline{0.714} & \textbf{20.24} \\
\addlinespace[2pt]
\rowcolor{mmegroupgray}
\multicolumn{6}{@{}l}{\textbf{Closed-source}} \\
\midrule
Vidu~\citep{bao2024vidu} & \textbf{0.195} & \textbf{0.786} & \underline{15.96} & \underline{0.714} & \textbf{20.05} \\
Kling~\citep{klingteam2026motioncontrol} & \underline{0.197} & \underline{0.777} & \textbf{16.41} & \textbf{0.721} & \underline{23.14} \\
\botrule
\end{tabular*}
\end{table*}

\begin{table*}[!t]
\caption{
Subjective results of the task of vision-driven video generation. Bold and underlined values indicate the best and second-best results.
% \czb{Subjective MOS results for vision-driven video generation}. 
% Bold and underlined values indicate the best and second-best results, respectively.
% Perceptual evaluation of vision-driven synthesis (1--5 MOS) across three comparison blocks: open-source Head Animation, open-source Body Animation, and closed-source Body Animation. Rank emphasis is computed separately within each block.
}
\label{tab:vision_driven_human_mos}
\centering
\footnotesize
\setlength{\tabcolsep}{5pt}
\renewcommand{\arraystretch}{0.96}
\begin{tabular*}{\textwidth}{@{\extracolsep\fill}>{\raggedright\arraybackslash}p{0.34\textwidth}>{\centering\arraybackslash}p{0.14\textwidth}*{2}{>{\centering\arraybackslash}p{0.22\textwidth}}@{}}
\toprule
Model & ID Similarity $\uparrow$ & Emotion Naturalness $\uparrow$ & Motion Naturalness $\uparrow$ \\
\midrule
\rowcolor{mmegroupblue}
\multicolumn{4}{@{}l}{\textbf{Open-source}} \\
\midrule
\multicolumn{4}{@{}l}{\textit{Head Animation}} \\
HunyuanPortrait~\citep{xu2025hunyuanportrait} & 4.15 & 3.55 & \underline{3.66} \\
X-NeMo~\citep{zhao2025xnemo} & \underline{4.16} & \textbf{3.68} & \textbf{3.82} \\
PersonaLive!~\citep{li2026personalive} & \textbf{4.41} & \underline{3.67} & 3.63 \\
AniPortrait~\citep{wei2024aniportrait} & 4.04 & 3.11 & 3.22 \\
\addlinespace[1pt]
\multicolumn{4}{@{}l}{\textit{Body Animation}} \\
MimicMotion~\citep{zhang2025mimicmotion} & 1.36 & 1.48 & 2.05 \\
StableAnimator~\citep{tu2025stableanimator} & 1.70 & 1.56 & \underline{2.21} \\
Animate-X~\citep{tan2025animatex} & \underline{1.90} & \underline{1.60} & 1.95 \\
Wan2.2~\citep{wanvideo2025wan22} & \textbf{4.82} & \textbf{4.61} & \textbf{4.72} \\
\addlinespace[2pt]
\rowcolor{mmegroupgray}
\multicolumn{4}{@{}l}{\textbf{Closed-source}} \\
\midrule
Vidu~\citep{bao2024vidu} & \underline{4.73} & \textbf{4.60} & \underline{4.61} \\
Kling~\citep{klingteam2026motioncontrol} & \textbf{4.76} & \underline{4.59} & \textbf{4.64} \\
\botrule
\end{tabular*}
\end{table*}

\textbf{Vision-driven generation}. Table~\ref{tab:vision_driven_synthesis} reports the vision-driven generation results across the two open-source paradigms and the closed-source group.
Among the Head Animation systems, X-NeMo leads LPIPS (0.520), PSNR (10.15), and FID (150.57), whereas AniPortrait leads CSIM (0.884) and SSIM (0.383), indicating that no single system dominates both the reconstruction and the identity criteria. 
Among the Body Animation systems, 
Animate-X leads LPIPS (0.139), PSNR (18.81), and SSIM (0.755), while Wan2.2 leads CSIM (0.783) and FID (20.24).
Within the closed-source group, Kling leads PSNR (16.41) and SSIM (0.721), whereas Vidu leads LPIPS (0.195), CSIM (0.786), and FID (20.05). 

The subjective study in Table~\ref{tab:vision_driven_human_mos} offers a complementary view that largely reinforces the automatic results.
Within the open-source Head Animation group, PersonaLive! attains the highest identity mean of 4.41, whereas X-NeMo leads both emotion naturalness (3.68) and motion naturalness (3.82), mirroring the automatic metrics in that identity preservation and motion quality separate across systems. 
Among the open-source Body Animation systems, Wan2.2 leads all three criteria, with means of 4.82 for identity similarity, 4.61 for emotion naturalness, and 4.72 for motion naturalness.
Within the closed-source group, Kling leads identity similarity and motion naturalness while Vidu leads emotion naturalness, echoing the split observed for the appearance-level metrics. 
Notably, although Wan2.2 is an open-source model, its subjective scores exceed those of both Vidu and Kling, demonstrating its superior performance on our dataset.

\begin{figure*}[!t]
\centering
\includegraphics[width=0.88\textwidth]{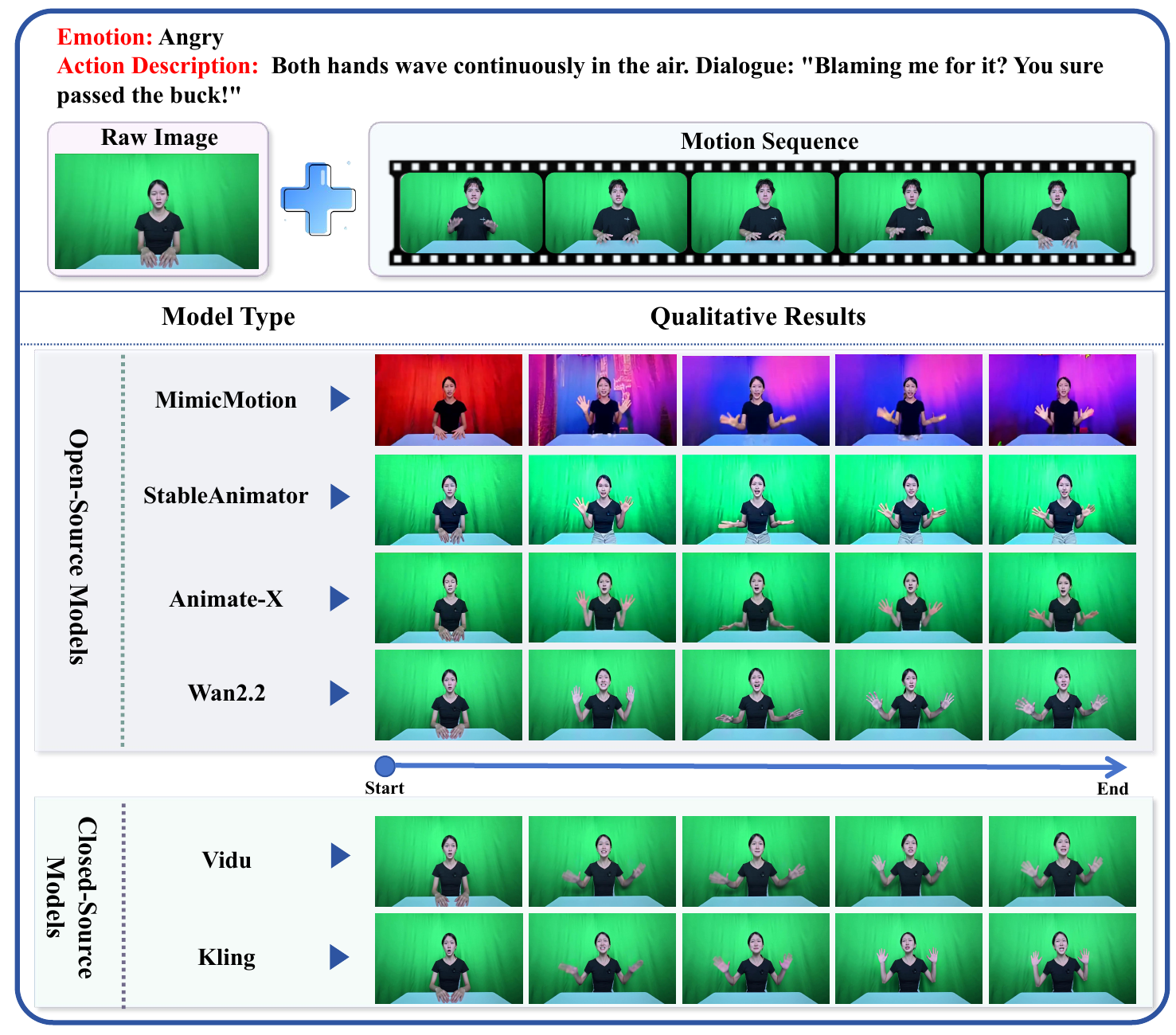}
\caption{
Qualitative comparison of vision-driven video generation for one Angry clip, with the source frame and driving motion sequence on top and temporally aligned outputs of each model below, grouped into open-source and closed-source.
}
\label{fig:temporal_synthesis_comparison}
\end{figure*}

\begin{figure*}[!t]
\centering
\includegraphics[width=0.88\textwidth]{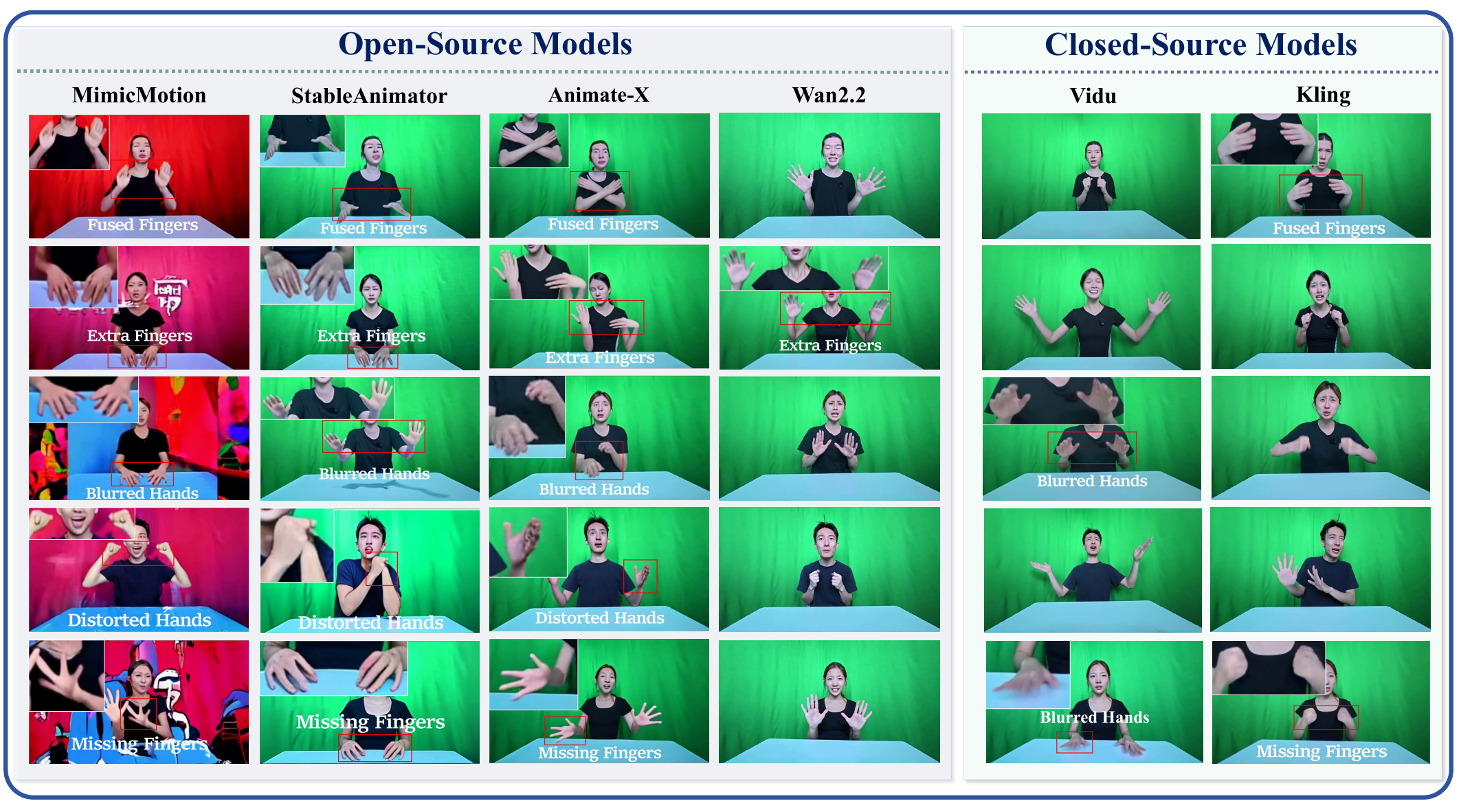}
\caption{
Fine-grained hand-region failures in vision-driven synthesis, highlighting fused, extra, blurred, distorted, and missing fingers or hands across six systems via enlarged insets. 
}
\label{fig:hand_artifact_comparison}
\end{figure*}

Moreover, the aligned sequence from the Angry clip in Fig.~\ref{fig:temporal_synthesis_comparison} shows that the models differ in when the arm trajectory develops, whether the requested gesture is ultimately completed, and whether appearance and background remain stable across the sampled sequence.
Fig.~\ref{fig:hand_artifact_comparison} isolates a complementary local axis by focusing on the hand region. Such structural errors occupy small image areas and are easily diluted by frame-averaged metrics. 
In these aspects, closed-source models are generally superior to open-source models.
Overall, 
these two figures demonstrate that a model may have a high average objective score but still exhibit deficiencies in the temporal or fine spatial dimensions, further highlighting the value of combining automated metrics with subjective analysis in this study.

\subsection{Voice Cloning}
\label{sec:voice_cloning_results}
Table~\ref{tab:voice_cloning} reports automatic voice-cloning results for six models, with real audio included as the speaker-similarity reference at 0.990.

The automatic measures capture two complementary properties: the reference-free quality predictors UTMOS and DNSMOS, and speaker similarity for identity preservation.
Among the open-source systems, OpenAudio S1 attains the highest UTMOS (2.32) and DNSMOS (3.01), whereas CosyVoice 3 achieves the highest speaker similarity (0.856).
Within the closed-source pair, Inworld TTS-1.5 leads UTMOS (2.69) and DNSMOS (3.22), while Eleven Multilingual v2 better preserves speaker identity (0.779).

The subjective analysis in Table~\ref{tab:voice_human_mos} provides another result: IndexTTS2 leads both open-source criteria, with a similarity mean of 4.24 and an emotion-naturalness mean of 4.40.
Within the closed-source group, Eleven Multilingual v2 obtains the higher means of 3.73 and 4.23 and surpasses Inworld TTS-1.5 despite the latter's advantage on the reference-free predictors.
Moreover, IndexTTS2 remains close to the best speaker similarity among the open-source models, whereas Eleven Multilingual v2 attains the highest speaker similarity among the closed-source models.
This suggests that the subjective-study leaders align more closely with speaker-identity preservation than with UTMOS or DNSMOS.
Conversely, the models that lead the reference-free predictors, OpenAudio S1 and Inworld TTS-1.5, do not lead subjectively.
Therefore, perceived voice-cloning quality appears to depend more strongly on speaker preservation than on the estimated signal, which helps explain the difference in ordering between Tables~\ref{tab:voice_cloning} and~\ref{tab:voice_human_mos}.

\begin{table*}[!t]
\caption{
Objective results of voice cloning. Bold and underlined values indicate the best and second-best results.
}\label{tab:voice_cloning}
\centering
\scriptsize
\setlength{\tabcolsep}{3pt}
\renewcommand{\arraystretch}{0.96}
\begin{tabular*}{\textwidth}{@{\extracolsep\fill}
>{\raggedright\arraybackslash}p{0.48\textwidth}
>{\centering\arraybackslash}p{0.13\textwidth}
>{\centering\arraybackslash}p{0.13\textwidth}
>{\centering\arraybackslash}p{0.16\textwidth}@{}}
\toprule
Model & UTMOS$\uparrow$ & DNSMOS$\uparrow$ & \mbox{Speaker Sim.$\uparrow$} \\
\midrule
\rowcolor{mmegroupblue}
\multicolumn{4}{@{}l}{\textbf{Reference}} \\
\midrule
Real audio & 1.90 & 2.84 & 0.990 \\
\addlinespace[2pt]
\rowcolor{mmegroupgray}
\multicolumn{4}{@{}l}{\textbf{Open-source models}} \\
\midrule
GPT-SoVITS v2~\citep{rvcboss2024gptsovitsv2} & 2.26 & 2.76 & 0.770 \\
IndexTTS2~\citep{zhou2026indextts2} & 2.00 & \underline{2.94} & 0.846 \\
CosyVoice 3~\citep{du2025cosyvoice3} & \underline{2.30} & \underline{2.94} & \textbf{0.856} \\
OpenAudio S1~\citep{openaudio2025s1} & \textbf{2.32} & \textbf{3.01} & \underline{0.848} \\
\addlinespace[2pt]
\rowcolor{mmegroupblue}
\multicolumn{4}{@{}l}{\textbf{Closed-source models}} \\
\midrule
Eleven Multilingual v2~\citep{elevenlabs2023multilingualv2} & \underline{2.12} & \underline{2.67} & \textbf{0.779} \\
Inworld TTS-1.5~\citep{inworld2026tts15} & \textbf{2.69} & \textbf{3.22} & \underline{0.767} \\
\botrule
\end{tabular*}
\end{table*}

\begin{table*}[!t]
\caption{
Subjective results for voice cloning. Bold and underlined values indicate the best and second-best results.
}
\label{tab:voice_human_mos}
\centering
\scriptsize
\setlength{\tabcolsep}{5pt}
\renewcommand{\arraystretch}{0.96}
\begin{tabular*}{\textwidth}{@{\extracolsep\fill}>{\raggedright\arraybackslash}p{0.43\textwidth}*{2}{>{\centering\arraybackslash}p{0.22\textwidth}}@{}}
\toprule
Model & ID Similarity $\uparrow$ & Emotion Naturalness $\uparrow$ \\
\midrule
\rowcolor{mmegroupblue}
\multicolumn{3}{@{}l}{\textbf{Open-source models}} \\
\midrule
GPT-SoVITS v2~\citep{rvcboss2024gptsovitsv2} & 2.99 & 3.12 \\
IndexTTS2~\citep{zhou2026indextts2} & \textbf{4.24} & \textbf{4.40} \\
CosyVoice 3~\citep{du2025cosyvoice3} & \underline{3.81} & \underline{4.29} \\
OpenAudio S1~\citep{openaudio2025s1} & 3.64 & 4.04 \\
\addlinespace[2pt]
\rowcolor{mmegroupgray}
\multicolumn{3}{@{}l}{\textbf{Closed-source models}} \\
\midrule
Eleven Multilingual v2~\citep{elevenlabs2023multilingualv2} & \textbf{3.73} & \textbf{4.23} \\
Inworld TTS-1.5~\citep{inworld2026tts15} & \underline{3.08} & \underline{3.67} \\
\botrule
\end{tabular*}
\end{table*}

\subsection{Human Video Matting}\label{sec:matting_results}

Table~\ref{tab:green_screen_matting}
reports the video-level performance of nine matting systems under the adapted green-screen protocol.
BiRefNet attains the best value on every criterion, reaching a MAD of 2.30, MSE of 0.82, dtSSD of 2.04, gradient error of 10.43, and connectivity error of 4.31. 
MatAnyone 2 ranks second throughout, with an MSE of 0.91 and dtSSD of 2.12 that trail BiRefNet only narrowly.
The stability of this ordering across both spatial and temporal criteria indicates that the leading methods couple accurate boundary estimation with reliable temporal propagation rather than improving one at the expense of the other.

\begin{figure*}[!t]
\centering
\includegraphics[width=0.88\textwidth]{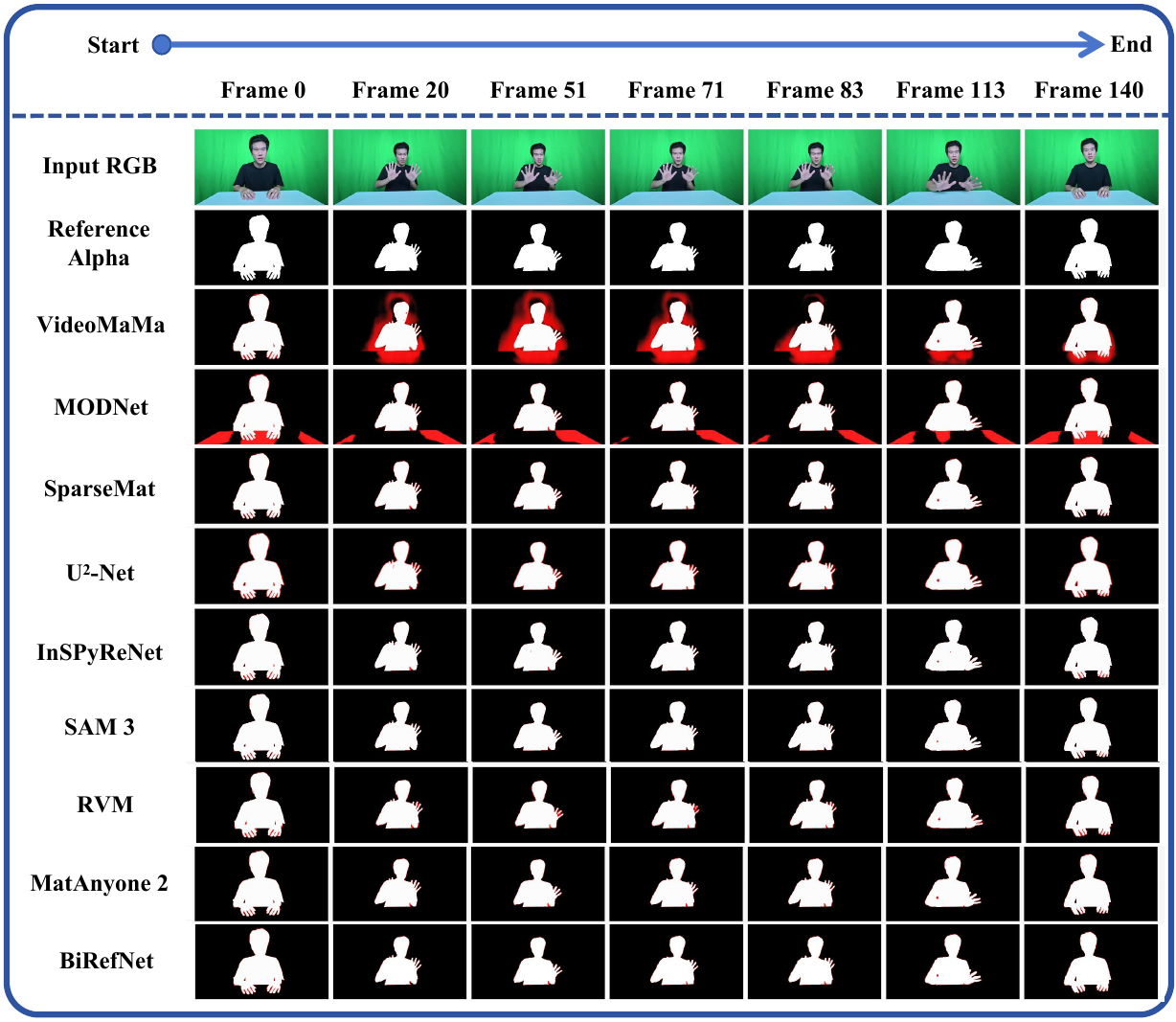}
\caption{
Qualitative comparison of human video matting on one green-screen sequence at seven time points, with rows showing the RGB input, the reference alpha, and the nine model predictions. 
Red visualization highlights boundary discrepancies relative to the reference alpha.
}
\label{fig:green_screen_matting_sequence}
\end{figure*}

Fig.~\ref{fig:green_screen_matting_sequence} localizes the residual errors and clarifies where they concentrate. 
The errors arise primarily from fine hand boundaries, transient motion contours, and foreground leakage. Even the most effective methods remain sensitive to slender fingers and rapidly changing hand shapes. 
These qualitative deficiencies are consistent with the quantitative gaps in gradient and connectivity error, which precisely penalize the boundary regions exposed by the sampled frames. Together, they indicate that boundary fidelity under motion is the primary challenge for human matting on this benchmark.

\begin{table*}[!t]
\caption{
Objective results of video matting.
Bold and underlined values indicate the best and second-best results.
}\label{tab:green_screen_matting}
\centering
\scriptsize
\setlength{\tabcolsep}{3pt}
\renewcommand{\arraystretch}{0.96}
\begin{tabular*}{\textwidth}{@{\extracolsep\fill}>{\raggedright\arraybackslash}p{0.42\textwidth}*{5}{>{\centering\arraybackslash}p{0.085\textwidth}}@{}}
\toprule
Model & MAD$\downarrow$ & MSE$\downarrow$ & dtSSD$\downarrow$ & Grad$\downarrow$ & Conn$\downarrow$ \\
\midrule
VideoMaMa~\citep{lim2026videomama} & 34.09 & 20.43 & 3.53 & 26.14 & 64.38 \\
MODNet~\citep{ke2020modnet} & 24.98 & 20.27 & 3.84 & 26.75 & 47.84 \\
SparseMat~\citep{sun2023sparsemat} & 9.14 & 4.27 & 3.19 & 28.83 & 14.79 \\
U$^2$-Net~\citep{qin2020u2net} & 6.78 & 2.49 & 2.76 & 33.11 & 10.33 \\
InSPyReNet~\citep{kim2022inspyrenet} & 4.33 & 2.52 & 3.58 & 31.79 & 5.77 \\
SAM 3~\citep{carion2025sam3} & 3.92 & 2.13 & 3.40 & 28.01 & 4.88 \\
RVM~\citep{lin2021rvm} & 4.80 & 1.65 & 2.38 & 16.55 & 6.64 \\
MatAnyone 2~\citep{yang2026matanyone2} & \underline{3.86} & \underline{0.91} & \underline{2.12} & \underline{11.45} & \underline{4.53} \\
BiRefNet~\citep{zheng2024birefnet} & \textbf{2.30} & \textbf{0.82} & \textbf{2.04} & \textbf{10.43} & \textbf{4.31} \\
\botrule
\end{tabular*}
\end{table*}

\begin{figure}[!t]
\centering
\includegraphics[width=\columnwidth]{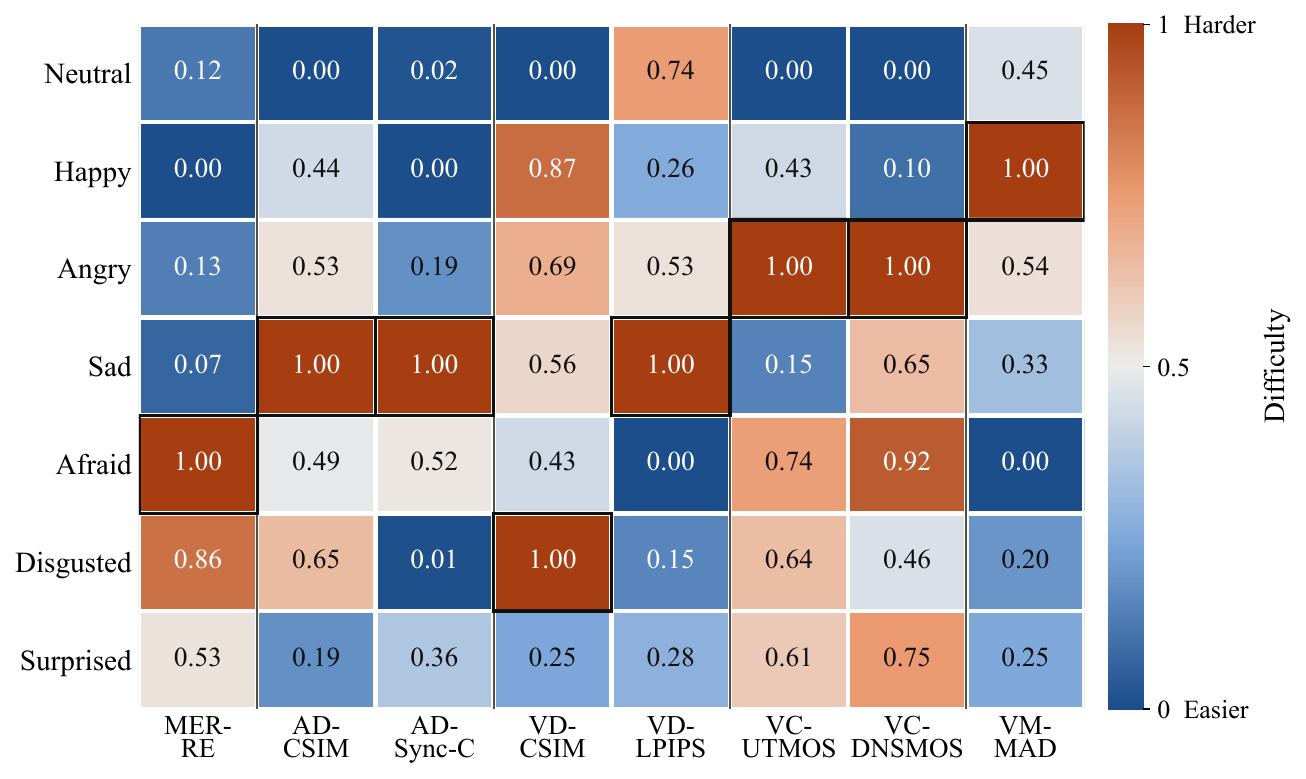}
\caption{
Emotion difficulty across eight metrics, each averaged over models, oriented so that larger values indicate greater difficulty, and independently min--max normalized over the seven emotions. Values of 0 and 1 denote the easiest and hardest conditions, respectively, and boxes mark column maxima.
}
\label{fig:emotion_metric_difficulty}
\end{figure}

\subsection{Cross-Task Analysis}
\label{sec:cross_task_analysis}

Following the setup in Section~\ref{subsec:cross_task_analysis_setup}, 
we present the cross-task analyses in the same order as their definitions.

\noindent\textbf{Emotion difficulty}.
We instantiate the emotion-by-metric difficulty with eight task-specific evaluation metrics, one or two per task, and we abbreviate each as a capability prefix followed by its underlying criterion. 
From multimodal emotion recognition, we take \textbf{MER-RE}, the recognition-error difficulty derived from seven-class accuracy. From audio-driven video generation, we take \textbf{AD-CSIM} (identity similarity) and \textbf{AD-Sync-C} (lip-synchronization confidence). From vision-driven video generation, we take \textbf{VD-CSIM} (identity similarity) and \textbf{VD-LPIPS} (perceptual reconstruction). From voice cloning, we take \textbf{VC-UTMOS} and \textbf{VC-DNSMOS}, the two reference-free speech-quality predictors. From video matting, we take \textbf{VM-MAD}, the mean absolute difference against the reference alpha. 
Then, these metrics are either retained, negated, or complemented to align with the direction of \(D_{e,k}\) in Eq.~(\ref{eq:cross_task_difficulty}) (larger values denote greater difficulty).

Fig.~\ref{fig:emotion_metric_difficulty} shows that the hardest condition depends strongly on both the capability and the criterion. 
Afraid is hardest for MER-RE, with Disgusted also difficult to recognize, consistent with the subtle or heterogeneous visual cues discussed in Section~\ref{sec:emotion_recognition_results}. 
Sad is hardest for AD-CSIM, AD-Sync-C, and VD-LPIPS, whereas Disgusted is hardest for VD-CSIM. 
Both voice-cloning dimensions, VC-UTMOS and VC-DNSMOS, identify Angry as the most difficult condition, since its forceful expressive delivery may be less favored by reference-free speech-quality predictors. 
Instead, VM-MAD peaks for Happy, whose expansive motion and open posture create larger foreground and boundary changes.

\begin{figure}[!t]
\centering
\includegraphics[width=\columnwidth]{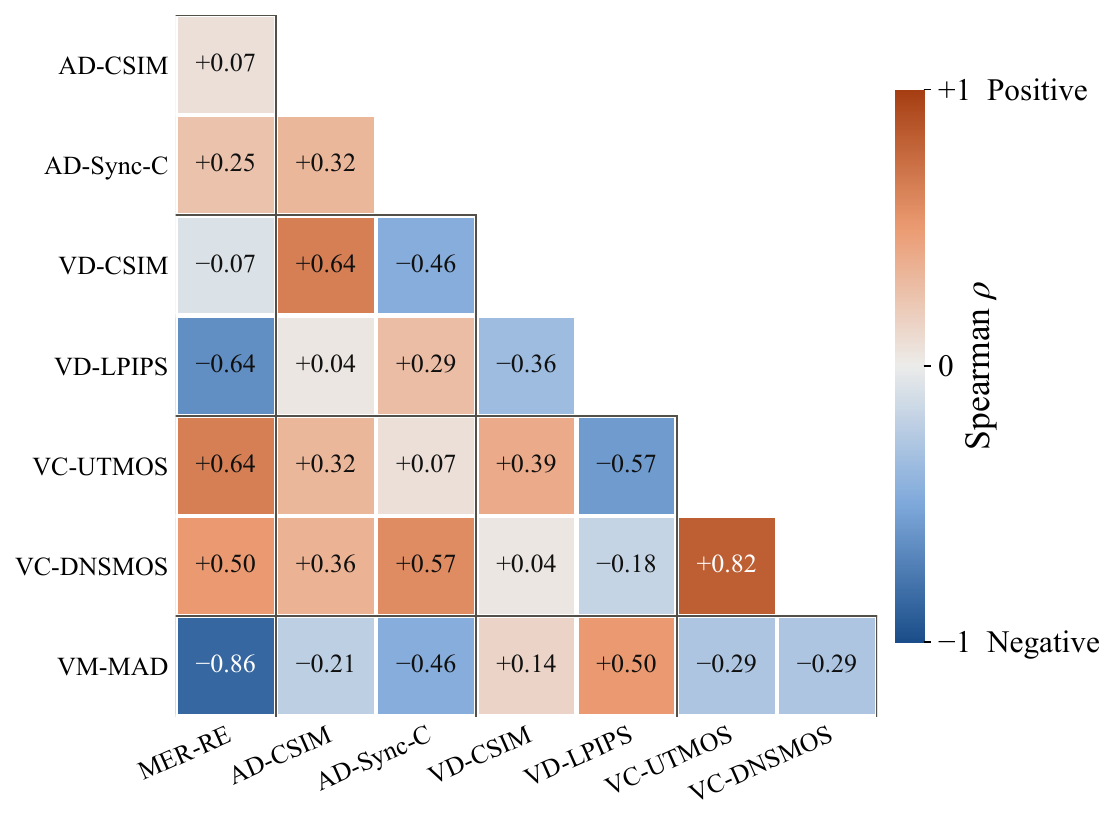}
\caption{
Pairwise Spearman rank correlations between the emotion difficulty profiles of the eight metrics. Positive and negative values indicate positive and negative correlations, respectively.
}
\label{fig:metric_profile_agreement}
\end{figure}

\noindent\textbf{Metric agreement}.
Applying the Spearman rank correlation of Eq.~(\ref{eq:cross_AGREEMENT}) to the emotion difficulty profiles of the eight metrics, 
Fig.~\ref{fig:metric_profile_agreement} shows that agreement is selective rather than universal. 
The two voice-quality profiles, VC-UTMOS and VC-DNSMOS, are strongly aligned (\(\rho = 0.82\)), and AD-CSIM and VD-CSIM share a broadly similar condition ordering, indicating within-family consistency. 
By contrast, MER-RE and VM-MAD are strongly opposed (\(\rho = -0.86\)), while most remaining pairs show weak or mixed agreement. 
These patterns reinforce that an emotion-level difficulty ranking must be read with respect to its capability and criterion rather than as a global property of the dataset.

\begin{figure}[!t]
\centering
\includegraphics[width=\columnwidth]{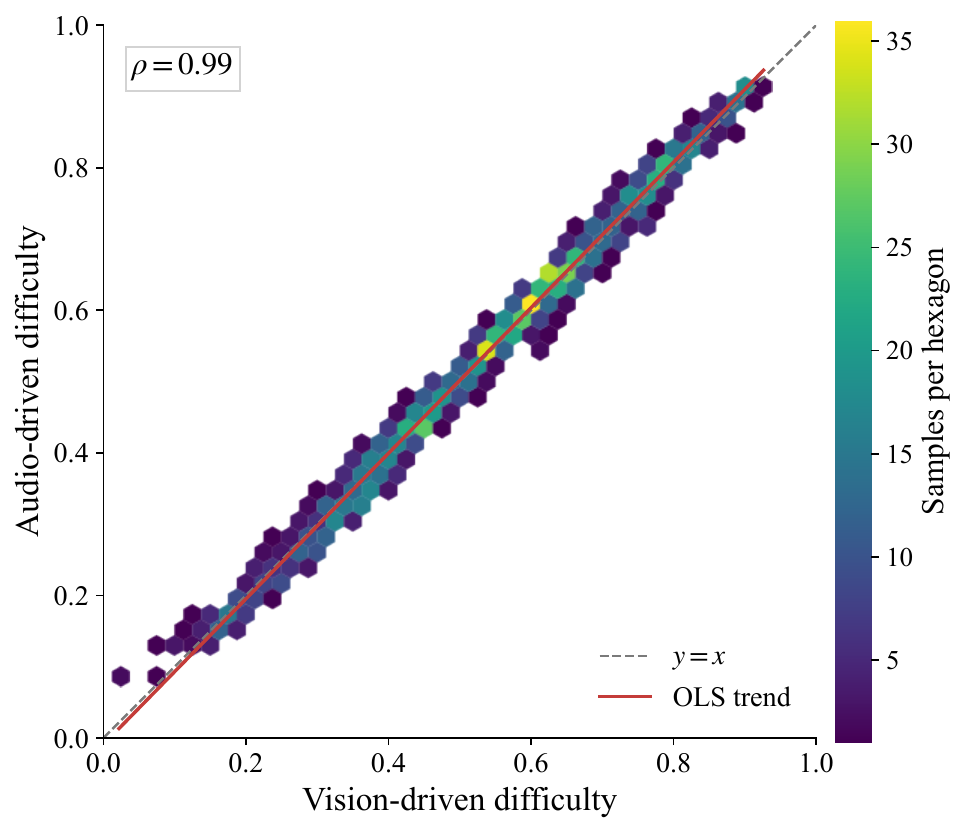}
\caption{
Source-level difficulty of the audio-driven and vision-driven methods on the the common set of valid clips.
Hexagon color encodes the number of samples per bin, the dashed line denotes equality, and the red line shows the ordinary least-squares (OLS) fit.
}
\label{fig:head_synthesis_agreement}
\end{figure}

\begin{figure}[!t]
\centering
\includegraphics[width=\columnwidth]{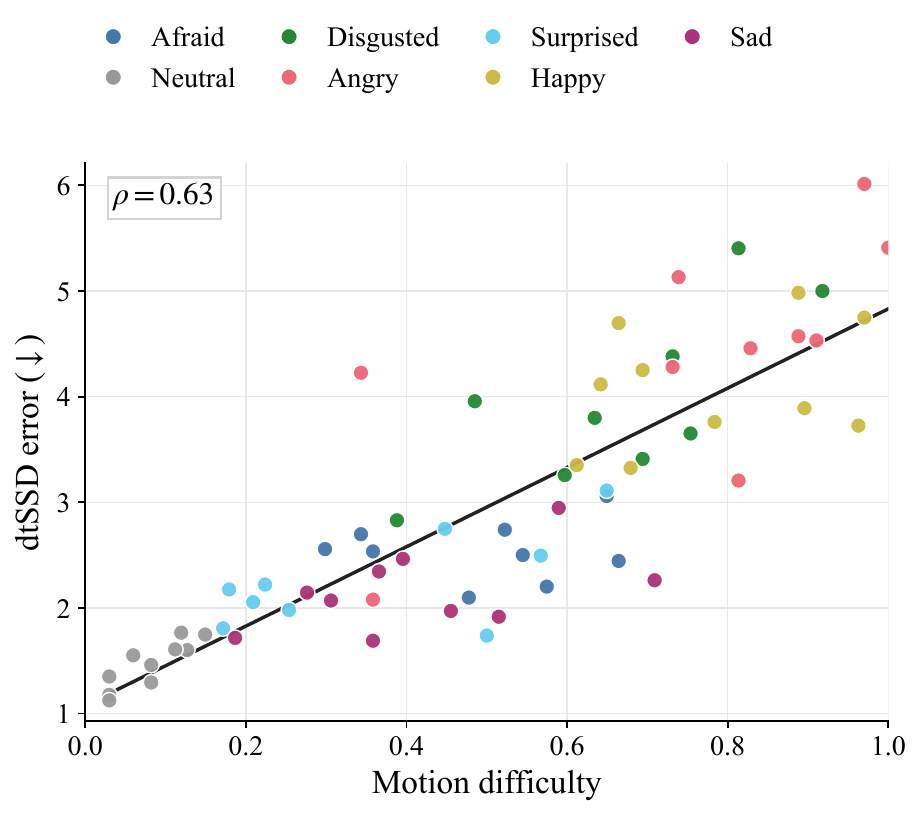}
\caption{
Relationship between motion difficulty and dtSSD, with the line denoting the ordinary least-squares (OLS) fit.
}
\label{fig:gt_difficulty_mechanisms}
\end{figure}

\noindent\textbf{Source-level difficulty}.
Following Eq.~(\ref{eq:cross_soure_level}), 
we aggregate the metric values within the audio-driven (AD) and vision-driven (VD) regimes and compare the resulting clip-level difficulty on the common-valid intersection.
Fig.~\ref{fig:head_synthesis_agreement} reveals nearly identical orderings for the two regimes (Spearman \(\rho = 0.99\)).
Although the two paradigms use different driving signals, their difficulty is built from the same appearance-fidelity and perceptual-reconstruction criteria on matched source clips. 
Therefore, the agreement reflects a shared sensitivity to the visual content and reconstruction demands of each source.

\noindent\textbf{Motion difficulty}.
Finally, based on Eq.~(\ref{eq:motion1}) and Eq.~(\ref{eq:motion2}), we analyze the motion difficulty \(D_{\text{motion}}(s)\) and its relationship with the dtSSD error.
Fig.~\ref{fig:gt_difficulty_mechanisms} shows a clear positive association (\(\rho = 0.63\)). 
Neutral samples cluster at low motion difficulty and low temporal error, consistent with the gentle, stable and low-amplitude actions specified in Table~\ref{tab:emotion_action_prompt_design}, whereas Happy, Angry, and Disgusted involve more expansive, forceful, or outward-moving actions and thus exhibit larger alpha and centroid changes together with higher dtSSD error. 
In addition, this trend echoes the temporal boundary artifacts reported in Section~\ref{sec:matting_results}, as larger foreground changes place greater demands on maintaining accurate alpha transitions over time.
Through this analysis, we characterize the motion difficulty across emotions effectively and validate the soundness of our dataset design.

\FloatBarrier

Overall, these analyses establish cross-task connections that cannot be achieved through single-task evaluation and yield important insights from quantitative comparisons, including (1) significant differences between emotion recognition and matting, (2) a consistent emphasis on generation quality in audio- and vision-driven video generation, (3) agreement between the two voice-cloning criteria, and (4) greater matting difficulty for emotions such as Happy, Angry, and Disgusted. They offer a useful preliminary attempt at jointly analyzing multiple understanding and generation tasks.

\section{Discussion}\label{sec:discussion}
Based on the above series of analyses, we distill deeper insights into how human-centered understanding and generation relate, and outline the future research directions.

\subsection{Insights}
The first insight concerns an asymmetry in how affect is handled across understanding and generation. 
In understanding, linguistic content is strongly predictive whereas visual evidence forms the weakest cue, in line with prior observations \citep{li2023decoupled,ma2025generative}. In generation, the burden shifts almost entirely onto pixels and motion, so that identity, expression, and gesture must be synthesized rather than inferred.
Consequently, current models can recognize emotions through linguistic shortcuts, but cannot rely solely on language when rendering emotions. 
This reveals a representational gap between how affect is read and how it is produced.
Closing this gap is central to human-centered intelligence, where meaning resides jointly in what is said and how it is enacted.

The second insight is that evaluation itself remains a bottleneck. In both generative tasks (video generation and voice cloning), task-specific quantitative and qualitative evaluations show a consistent overall trend, but diverge in the top rankings. Furthermore, 
our survey has revealed a lack of metrics for cross-task analysis, which motivates us to design a suite of such metrics that build upon task-specific measures and incorporate statistical analysis, as presented in Section~\ref{subsec:cross_task_analysis_setup}. 
The cross-task analyses in Section~\ref{sec:cross_task_analysis} show that their outcomes accord with prior empirical intuitions, thereby confirming their soundness.
Therefore, to further advance the research on understanding and generation, it is necessary to refine both task-specific and cross-task evaluation metrics.

The third insight is that cross-task analyses of understanding and generation within a coordinated system can offer valuable guidance for the emerging frontier of unified multimodal understanding and generation models \citep{tang2025humancentricfm,wang2026emu3}.
These analyses show that difficulty is relational rather than intrinsic, arising jointly from the emotion, the model, and the evaluation criterion. 
By quantifying how errors correlate or oppose across tasks, such as the negative coupling between recognition and matting and the close agreement between the two voice-quality predictors, these metrics reveal which capabilities share representational demands and which compete \citep{wu2025janus}.
This provides a principled basis for deciding how perception and synthesis should be coupled within a single architecture.

\subsection{Outlook}
These insights point to three promising directions. 
First, understanding calls for effective models and fusion mechanisms that better exploit visual cues, turning them from the weak link into a reliable source of affective evidence. Second, evaluation should couple automatic metrics with MOS so that the information can be jointly integrated into more principled measures, while extending such measures further across tasks. 
Third, 
multi-task architectures that explicitly regularize these correlations can treat the constituent understanding and generation tasks as mutually constraining objectives, thereby transforming the current benchmark from a diagnostic tool into a training substrate for more coherent human-centered visual intelligence systems.

\section{Conclusion}\label{sec:conclusion}
We present HUG-VIS, the first unified multimodal benchmark that places human-centered understanding and generation tasks on a shared actor-by-assignment grid.
The dataset contains 8,400 seated half-body performances from 30 actors on an identical 280-assignment grid, with synchronized video, audio, text, and alpha mattes. 
A series of analyses yields four consistent findings: (1) linguistic content dominates present-day emotion recognition while purely visual affect remains the weakest setting, (2) automatic metrics and human judgment agree in overall trend yet diverge at the top, so the two must be reported jointly for both generation tasks, (3) boundary fidelity under motion is the principal obstacle for human matting, and (4) task difficulty is not an intrinsic property of a condition but emerges jointly from the emotion, the capability, and the criterion.
By enabling understanding and generation on a shared and aligned foundation, we anticipate that HUG-VIS will serve as an important reference for human-centered visual intelligence.

\backmatter

\section*{Statements and Declarations}

\noindent\textbf{Consent to participate.}
Before recording, all actors provided written informed consent for research capture and for the authorized use of their identifiable likeness, voice, and performed behavior.

\noindent\textbf{Consent for publication.}
Publication and demonstration of identifiable examples are allowed for research purposes covered by the participating actors’ written authorizations.

\noindent\textbf{Data availability.}
The dataset, including its synchronized audio, video, text, and alpha assets, is released through the project page at \url{https://github.com/GML-MMGroup/HUG-VIS}.
Access is granted after application review and is limited to non-commercial academic research under the recipient data-use agreement.

\noindent\textbf{Code availability.}
The associated evaluation code is released through the same project page.

\noindent\textbf{Competing interests.}
The authors declare that they have no competing interests.

\bibliography{sn-bibliography}% common bib file

@article{busso2008iemocap,
  author = {Busso, Carlos and Bulut, Murtaza and Lee, Chi-Chun and Kazemzadeh, Abe and Mower, Emily and Kim, Samuel and Chang, Jeannette N. and Lee, Sungbok and Narayanan, Shrikanth S.},
  title = {{IEMOCAP}: Interactive Emotional Dyadic Motion Capture Database},
  journal = {Lang. Resour. Eval.},
  volume = {42},
  number = {4},
  pages = {335--359},
  year = {2008},
}

@article{baltrusaitis2019multimodal,
  author = {Baltru{\v{s}}aitis, Tadas and Ahuja, Chaitanya and Morency, Louis-Philippe},
  title = {Multimodal Machine Learning: A Survey and Taxonomy},
  journal = {IEEE TPAMI},
  volume = {41},
  number = {2},
  pages = {423--443},
  year = {2019},
}

@article{ma2025generative,
  title={Generative technology for human emotion recognition: A scoping review},
  author={Ma, Fei and Yuan, Yucheng and Xie, Yifan and Ren, Hongwei and Liu, Ivan and He, Ying and Ren, Fuji and Yu, Fei Richard and Ni, Shiguang},
  journal={Information Fusion},
  volume={115},
  pages={102753},
  year={2025},
  publisher={Elsevier}
}

@article{li2024multimodalfoundation,
  author = {Li, Chunyuan and Gan, Zhe and Yang, Zhengyuan and Yang, Jianwei and Li, Linjie and Wang, Lijuan and Gao, Jianfeng},
  title = {Multimodal Foundation Models: From Specialists to General-Purpose Assistants},
  journal = {Found. Trends Comput. Graph. Vis.},
  volume = {16},
  number = {1--2},
  pages = {1--214},
  year = {2024},
}

@inproceedings{ramesh2021zeroshot,
  author = {Ramesh, Aditya and Pavlov, Mikhail and Goh, Gabriel and Gray, Scott and Voss, Chelsea and Radford, Alec and Chen, Mark and Sutskever, Ilya},
  title = {Zero-Shot Text-to-Image Generation},
  booktitle = {{ICML}},
  volume = {139},
  pages = {8821--8831},
  year = {2021},
}

@inproceedings{ci2023unihcp,
  author = {Ci, Yuanzheng and Wang, Yizhou and Chen, Meilin and Tang, Shixiang and Bai, Lei and Zhu, Feng and Zhao, Rui and Yu, Fengwei and Qi, Donglian and Ouyang, Wanli},
  title = {{UniHCP}: A Unified Model for Human-Centric Perceptions},
  booktitle = {{CVPR}},
  pages = {17840--17852},
  year = {2023},
}

@inproceedings{tang2025humancentricfm,
  author = {Tang, Shixiang and Wang, Yizhou and Chen, Lu and Wang, Yuan and Peng, Sida and Xu, Dan and Ouyang, Wanli},
  title = {Human-Centric Foundation Models: Perception, Generation and Agentic Modeling},
  booktitle = {{IJCAI}},
  pages = {10678--10686},
  year = {2025},
}

@inproceedings{zhou2026humanvbench,
  author = {Zhou, Ting and Chen, Daoyuan and Jiao, Qirui and Ding, Bolin and Li, Yaliang and Shen, Ying},
  title = {{HumanVBench}: Probing Human-Centric Video Understanding in {MLLMs} with Automatically Synthesized Benchmarks},
  booktitle = {{CVPR}},
  pages = {4494--4504},
  year = {2026}
}

@inproceedings{chung2016out,
  author = {Chung, Joon Son and Zisserman, Andrew},
  title = {Out of Time: Automated Lip Sync in the Wild},
  booktitle = {{ACCV Workshops}},
  pages = {251--263},
  year = {2016},
}

@inproceedings{li2023decoupled,
  author = {Li, Yong and Wang, Yuanzhi and Cui, Zhen},
  title = {Decoupled Multimodal Distilling for Emotion Recognition},
  booktitle = {{CVPR}},
  pages = {6631--6640},
  year = {2023}
}

@article{mago2026abstract,
  author = {Mago, Gowreesh and Mettes, Pascal and Rudinac, Stevan},
  title = {Looking Beyond the Obvious: A Survey on Abstract Concept Recognition for Video Understanding},
  journal = {IJCV},
  volume = {134},
  pages = {217},
  year = {2026},
}

@article{xing2026emollama,
  author = {Xing, Bohao and Yu, Zitong and Liu, Xin and Yuan, Kaishen and Ye, Qilang and Xie, Weicheng and Yue, Huanjing and Yang, Jingyu and K{\"a}lvi{\"a}inen, Heikki},
  title = {{EMO-LLaMA}: Enhancing Facial Emotion Understanding with Instruction Tuning},
  journal = {IJCV},
  volume = {134},
  pages = {350},
  year = {2026},
}

@article{vougioukas2020speech,
  author = {Vougioukas, Konstantinos and Petridis, Stavros and Pantic, Maja},
  title = {Realistic Speech-Driven Facial Animation with {GANs}},
  journal = {IJCV},
  volume = {128},
  number = {5},
  pages = {1398--1413},
  year = {2020},
}

@article{bounareli2024oneshot,
  author = {Bounareli, Stella and Tzelepis, Christos and Argyriou, Vasileios and Patras, Ioannis and Tzimiropoulos, Georgios},
  title = {One-Shot Neural Face Reenactment via Finding Directions in {GAN}'s Latent Space},
  journal = {IJCV},
  volume = {132},
  number = {8},
  pages = {3324--3354},
  year = {2024},
}

@article{stergiou2025abouttime,
  author = {Stergiou, Alexandros and Poppe, Ronald},
  title = {About Time: Advances, Challenges, and Outlooks of Action Understanding},
  journal = {IJCV},
  volume = {133},
  number = {9},
  pages = {6251--6315},
  year = {2025},
}

@article{georgakis2018dynamic,
  author = {Georgakis, Christos and Panagakis, Yannis and Pantic, Maja},
  title = {Dynamic Behavior Analysis via Structured Rank Minimization},
  journal = {IJCV},
  volume = {126},
  pages = {333--357},
  year = {2018},
}

@article{wang2026emu3,
  author = {Wang, Xinlong and Cui, Yufeng and Wang, Jinsheng and Zhang, Fan and Wang, Yueze and Zhang, Xiaosong and Luo, Zhengxiong and Sun, Quan and Li, Zhen and Wang, Yuqi and Yu, Qiying and Zhao, Yingli and Ao, Yulong and Min, Xuebin and Men, Chunlei and Wu, Boya and Zhao, Bo and Zhang, Bowen and Wang, Liangdong and Liu, Guang and He, Zheqi and Yang, Xi and Liu, Jingjing and Lin, Yonghua and Wang, Zhongyuan and Huang, Tiejun},
  title = {Multimodal Learning with Next-Token Prediction for Large Multimodal Models},
  journal = {Nature},
  volume = {650},
  pages = {327--333},
  year = {2026},
}

@inproceedings{wu2025janus,
  author = {Wu, Chengyue and Chen, Xiaokang and Wu, Zhiyu and Ma, Yiyang and Liu, Xingchao and Pan, Zizheng and Liu, Wen and Xie, Zhenda and Yu, Xingkai and Ruan, Chong and Luo, Ping},
  title = {{Janus}: Decoupling Visual Encoding for Unified Multimodal Understanding and Generation},
  booktitle = {{CVPR}},
  pages = {12966--12977},
  year = {2025}
}

@inproceedings{ju2024naturalspeech,
  author = {Ju, Zeqian and Wang, Yuancheng and Shen, Kai and Tan, Xu and Xin, Detai and Yang, Dongchao and Liu, Eric and Leng, Yichong and Song, Kaitao and Tang, Siliang and Wu, Zhizheng and Qin, Tao and Li, Xiangyang and Ye, Wei and Zhang, Shikun and Bian, Jiang and He, Lei and Li, Jinyu and Zhao, Sheng},
  title = {{NaturalSpeech} 3: Zero-Shot Speech Synthesis with Factorized Codec and Diffusion Models},
  booktitle = {{ICML}},
  volume = {235},
  pages = {22605--22623},
  year = {2024},
}

@misc{yamagishi2019vctk,
  author = {Yamagishi, Junichi and Veaux, Christophe and MacDonald, Kirsten},
  title = {{CSTR VCTK Corpus}: English Multi-speaker Corpus for {CSTR} Voice Cloning Toolkit (version 0.92)},
  year = {2019},
}

@inproceedings{ke2020modnet,
  author = {Ke, Zhanghan and Sun, Jiayu and Li, Kaican and Yan, Qiong and Lau, Rynson W. H.},
  title = {{MODNet}: Real-Time Trimap-Free Portrait Matting via Objective Decomposition},
  booktitle = {{AAAI}},
  volume = {36},
  number = {1},
  pages = {1140--1147},
  year = {2022},
}

@inproceedings{lin2021bgmv2,
  author = {Lin, Shanchuan and Ryabtsev, Andrey and Sengupta, Soumyadip and Curless, Brian and Seitz, Steven M. and Kemelmacher-Shlizerman, Ira},
  title = {Real-Time High-Resolution Background Matting},
  booktitle = {{CVPR}},
  pages = {8762--8771},
  year = {2021}
}

@inproceedings{lin2021rvm,
  author = {Lin, Shanchuan and Yang, Linjie and Saleemi, Imran and Sengupta, Soumyadip},
  title = {Robust High-Resolution Video Matting with Temporal Guidance},
  booktitle = {{WACV}},
  pages = {238--247},
  year = {2022}
}

@inproceedings{liu2022beat,
  author = {Liu, Haiyang and Zhu, Zihao and Iwamoto, Naoya and Peng, Yichen and Li, Zhengqing and Zhou, You and Bozkurt, Elif and Zheng, Bo},
  title = {{BEAT}: A Large-Scale Semantic and Emotional Multi-Modal Dataset for Conversational Gestures Synthesis},
  booktitle = {{ECCV}},
  pages = {612--630},
  year = {2022}
}

@inproceedings{prajwal2020wav2lip,
  author = {Prajwal, K. R. and Mukhopadhyay, Rudrabha and Namboodiri, Vinay P. and Jawahar, C. V.},
  title = {A Lip Sync Expert Is All You Need for Speech to Lip Generation in the Wild},
  booktitle = {{ACM MM}},
  pages = {484--492},
  year = {2020},
}

@inproceedings{reddy2021dnsmos,
  author = {Reddy, Chandan K. A. and Gopal, Vishak and Cutler, Ross},
  title = {{DNSMOS}: A Non-Intrusive Perceptual Objective Speech Quality Metric to Evaluate Noise Suppressors},
  booktitle = {{ICASSP}},
  pages = {6493--6497},
  year = {2021}
}

@inproceedings{saeki2022utmos,
  author = {Saeki, Takaaki and Xin, Detai and Nakata, Wataru and Koriyama, Tomoki and Takamichi, Shinnosuke and Saruwatari, Hiroshi},
  title = {{UTMOS}: {UTokyo-SaruLab} System for {VoiceMOS} Challenge 2022},
  booktitle = {{INTERSPEECH}},
  pages = {4521--4525},
  year = {2022},
}

@incollection{wang2020mead,
  author = {Wang, Kaisiyuan and Wu, Qianyi and Song, Linsen and Yang, Zhuoqian and Wu, Wayne and Qian, Chen and He, Ran and Qiao, Yu and Loy, Chen Change},
  title = {{MEAD}: A Large-Scale Audio-Visual Dataset for Emotional Talking-Face Generation},
  booktitle = {{ECCV}},
  pages = {700--717},
  year = {2020}
}

@inproceedings{wan2018generalized,
  author = {Wan, Li and Wang, Quan and Papir, Alan and Lopez Moreno, Ignacio},
  title = {Generalized End-to-End Loss for Speaker Verification},
  booktitle = {{ICASSP}},
  pages = {4879--4883},
  year = {2018}
}

@inproceedings{yang2025matanyone,
  author = {Yang, Peiqing and Zhou, Shangchen and Zhao, Jixin and Tao, Qingyi and Loy, Chen Change},
  title = {{MatAnyone}: Stable Video Matting with Consistent Memory Propagation},
  booktitle = {{CVPR}},
  pages = {7299--7308},
  year = {2025}
}

@inproceedings{zhang2021hdtf,
  author = {Zhang, Zhimeng and Li, Lincheng and Ding, Yu and Fan, Changjie},
  title = {Flow-Guided One-Shot Talking Face Generation With a High-Resolution Audio-Visual Dataset},
  booktitle = {{CVPR}},
  pages = {3661--3670},
  year = {2021}
}

@inproceedings{zadeh2018mosei,
  author = {Zadeh, AmirAli Bagher and Liang, Paul Pu and Poria, Soujanya and Cambria, Erik and Morency, Louis-Philippe},
  title = {Multimodal Language Analysis in the Wild: {CMU-MOSEI} Dataset and Interpretable Dynamic Fusion Graph},
  booktitle = {{ACL}},
  pages = {2236--2246},
  year = {2018}
}

@article{ekman1992basic,
  author = {Ekman, Paul},
  title = {An Argument for Basic Emotions},
  journal = {Cogn. Emot.},
  volume = {6},
  number = {3--4},
  pages = {169--200},
  year = {1992}
}

@article{demeijer1989movement,
  author = {de Meijer, Marco},
  title = {The Contribution of General Features of Body Movement to the Attribution of Emotions},
  journal = {J. Nonverbal Behav.},
  volume = {13},
  number = {4},
  pages = {247--268},
  year = {1989}
}

@article{wallbott1998bodily,
  author = {Wallbott, Harald G.},
  title = {Bodily Expression of Emotion},
  journal = {Eur. J. Soc. Psychol.},
  volume = {28},
  number = {6},
  pages = {879--896},
  year = {1998}
}

@article{atkinson2004body,
  author = {Atkinson, Anthony P. and Dittrich, Winand H. and Gemmell, Andrew J. and Young, Andrew W.},
  title = {Emotion Perception from Dynamic and Static Body Expressions in Point-Light and Full-Light Displays},
  journal = {Perception},
  volume = {33},
  number = {6},
  pages = {717--746},
  year = {2004}
}

@article{frijda1989action,
  author = {Frijda, Nico H. and Kuipers, Peter and ter Schure, Elisabeth},
  title = {Relations among Emotion, Appraisal, and Emotional Action Readiness},
  journal = {J. Pers. Soc. Psychol.},
  volume = {57},
  number = {2},
  pages = {212--228},
  year = {1989}
}

@article{banziger2012gemep,
  author = {B{\"a}nziger, Tanja and Mortillaro, Marcello and Scherer, Klaus R.},
  title = {Introducing the {Geneva Multimodal Expression Corpus} for Experimental Research on Emotion Perception},
  journal = {Emotion},
  volume = {12},
  number = {5},
  pages = {1161--1179},
  year = {2012}
}

@inproceedings{zhang2018lpips,
  author = {Zhang, Richard and Isola, Phillip and Efros, Alexei A. and Shechtman, Eli and Wang, Oliver},
  title = {The Unreasonable Effectiveness of Deep Features as a Perceptual Metric},
  booktitle = {{CVPR}},
  pages = {586--595},
  year = {2018},
}

@inproceedings{heusel2017fid,
  author = {Heusel, Martin and Ramsauer, Hubert and Unterthiner, Thomas and Nessler, Bernhard and Hochreiter, Sepp},
  title = {{GANs} Trained by a Two Time-Scale Update Rule Converge to a Local {Nash} Equilibrium},
  booktitle = {{NeurIPS}},
  volume = {30},
  year = {2017}
}

@inproceedings{erofeev2015video,
  author = {Erofeev, Mikhail and Gitman, Yury and Vatolin, Dmitriy and Fedorov, Alexey and Wang, Jue},
  title = {Perceptually Motivated Benchmark for Video Matting},
  booktitle = {{BMVC}},
  pages = {99.1--99.12},
  year = {2015},
}

@article{wang2004ssim,
  author = {Wang, Zhou and Bovik, Alan C. and Sheikh, Hamid R. and Simoncelli, Eero P.},
  title = {Image Quality Assessment: From Error Visibility to Structural Similarity},
  journal = {IEEE TIP},
  volume = {13},
  number = {4},
  pages = {600--612},
  year = {2004},
}

@inproceedings{deng2019arcface,
  author = {Deng, Jiankang and Guo, Jia and Xue, Niannan and Zafeiriou, Stefanos},
  title = {{ArcFace}: Additive Angular Margin Loss for Deep Face Recognition},
  booktitle = {{CVPR}},
  pages = {4690--4699},
  year = {2019},
}

@inproceedings{zhao2021formerdfer,
  author = {Zhao, Zengqun and Liu, Qingshan},
  title = {{Former-DFER}: Dynamic Facial Expression Recognition Transformer},
  booktitle = {{ACM MM}},
  pages = {1553--1561},
  year = {2021},
}

@inproceedings{meng2019fan,
  author = {Meng, Debin and Peng, Xiaojiang and Wang, Kai and Qiao, Yu},
  title = {Frame Attention Networks for Facial Expression Recognition in Videos},
  booktitle = {{ICIP}},
  pages = {3866--3870},
  year = {2019},
}

@misc{emotiefflib2025,
  author = {{Sber AI Lab}},
  title = {{EmotiEffLib}: Efficient Face Emotion Recognition in Photos and Videos},
  year = {2025},
}

@inproceedings{savchenko2023adaptive,
  author = {Savchenko, Andrey},
  title = {Facial Expression Recognition with Adaptive Frame Rate Based on Multiple Testing Correction},
  booktitle = {{ICML}},
  volume = {202},
  pages = {30119--30129},
  year = {2023},
}

@inproceedings{chumachenko2024mmadfer,
  author = {Chumachenko, Kateryna and Iosifidis, Alexandros and Gabbouj, Moncef},
  title = {{MMA-DFER}: Multimodal Adaptation of Unimodal Models for Dynamic Facial Expression Recognition In-the-Wild},
  booktitle = {{CVPR Workshops}},
  pages = {4673--4682},
  year = {2024}
}

@article{qwen2025qwen25omni,
  author = {Xu, Jin and Guo, Zhifang and He, Jinzheng and Hu, Hangrui and He, Ting and Bai, Shuai and Chen, Keqin and Wang, Jialin and Fan, Yang and Dang, Kai and Zhang, Bin and Wang, Xiong and Chu, Yunfei and Lin, Junyang},
  title = {{Qwen2.5-Omni} Technical Report},
  journal = {arXiv preprint arXiv:2503.20215},
  year = {2025},
}

@article{zhao2025humanomni,
  author = {Zhao, Jiaxing and Yang, Qize and Peng, Yixing and Bai, Detao and Yao, Shimin and Sun, Boyuan and Chen, Xiang and Fu, Shenghao and Chen, Weixuan and Wei, Xihan and Bo, Liefeng},
  title = {{HumanOmni}: A Large Vision-Speech Language Model for Human-Centric Video Understanding},
  journal = {arXiv preprint arXiv:2501.15111},
  year = {2025},
}

@article{zhang2025videollama3,
  author = {Zhang, Boqiang and Li, Kehan and Cheng, Zesen and Hu, Zhiqiang and Yuan, Yuqian and Chen, Guanzheng and Leng, Sicong and Jiang, Yuming and Zhang, Hang and Li, Xin and Jin, Peng and Zhang, Wenqi and Wang, Fan and Bing, Lidong and Zhao, Deli},
  title = {{VideoLLaMA 3}: Frontier Multimodal Foundation Models for Image and Video Understanding},
  journal = {arXiv preprint arXiv:2501.13106},
  year = {2025},
}

@inproceedings{yu2025minicpmv45,
  author = {Yu, Tianyu and Wang, Zefan and Wang, Chongyi and Huang, Fuwei and Ma, Wenshuo and He, Zhihui and Cai, Tianchi and Chen, Weize and Huang, Yuxiang and Zhao, Ranchi and Xu, Bokai and Cui, Junbo and Xu, Yingjing and Ruan, Liqing and Zhang, Luoyuan and Liu, Hanyu and Tang, Jingkun and Liu, Hongyuan and Guo, Qining and Hu, Wenhao and He, Bingxiang and Zhou, Jie and Cai, Jie and Qi, Ji and Guo, Zonghao and Chen, Chi and Zeng, Guoyang and Li, Yuxuan and Cui, Ganqu and Ding, Ning and Han, Xu and Yao, Yuan and Liu, Zhiyuan and Sun, Maosong},
  title = {{MiniCPM-V 4.5}: Cooking Efficient {MLLMs} via Architecture, Data, and Training Recipe},
  booktitle = {{CVPR}},
  pages = {11704--11715},
  year = {2026},
}

@inproceedings{clark2026molmo2,
  author = {Clark, Christopher and Zhang, Jieyu and Ma, Zixian and Park, Jae Sung and Tripathi, Rohun and Lee, Sangho and Salehi, Mohammadreza and Ren, Jason and Kim, Chris Dongjoo and Yang, Yinuo and Shao, Vincent and Yang, Yue and Huang, Weikai and Gao, Ziqi and Anderson, Taira and Zhang, Jianrui and Jain, Jitesh and Stoica, George and Farhadi, Ali and Krishna, Ranjay},
  title = {{Molmo2}: Open Weights and Data for Vision-Language Models with Video Understanding and Grounding},
  booktitle = {{CVPR}},
  pages = {28652--28668},
  year = {2026},
}

@article{wang2025internvideo25,
  author = {Wang, Yi and Li, Xinhao and Yan, Ziang and He, Yinan and Yu, Jiashuo and Zeng, Xiangyu and Wang, Chenting and Ma, Changlian and Huang, Haian and Gao, Jianfei and Dou, Min and Chen, Kai and Wang, Wenhai and Qiao, Yu and Wang, Yali and Wang, Limin},
  title = {{InternVideo2.5}: Empowering Video {MLLMs} with Long and Rich Context Modeling},
  journal = {arXiv preprint arXiv:2501.12386},
  year = {2025},
}

@article{cui2026minicpmo45,
  author = {Cui, Junbo and Xu, Bokai and Wang, Chongyi and Yu, Tianyu and Sun, Weiyue and Xu, Yingjing and Wang, Tianran and He, Zhihui and Ma, Wenshuo and Cai, Tianchi and Gui, Jiancheng and Zhang, Luoyuan and Sun, Xian and Huang, Fuwei and Chen, Moye and Lin, Zhuo and Liu, Hanyu and Gui, Qingxin and Han, Qingzhe and Wen, Yuyang and Liu, Huiping and Wang, Rongkang and Zhang, Yaqi and Wei, Hongliang and Chen, Chi and Li, You and Fang, Kechen and Zhou, Jie and Li, Yuxuan and Zeng, Guoyang and Xiao, Chaojun and Lin, Yankai and Han, Xu and Sun, Maosong and Liu, Zhiyuan and Yao, Yuan},
  title = {{MiniCPM-o 4.5}: Towards Real-Time Full-Duplex Omni-Modal Interaction},
  journal = {arXiv preprint arXiv:2604.27393},
  year = {2026},
}

@article{kimiteam2025kimiaudio,
  author = {{Kimi Team}},
  title = {{Kimi-Audio} Technical Report},
  journal = {arXiv preprint arXiv:2504.18425},
  year = {2025},
}

@article{chu2024qwen2audio,
  author = {Chu, Yunfei and Xu, Jin and Yang, Qian and Wei, Haojie and Wei, Xipin and Guo, Zhifang and Leng, Yichong and Lv, Yuanjun and He, Jinzheng and Lin, Junyang and Zhou, Chang and Zhou, Jingren},
  title = {{Qwen2-Audio} Technical Report},
  journal = {arXiv preprint arXiv:2407.10759},
  year = {2024},
}

@inproceedings{wang2026emotionthinker,
  author = {Wang, Dingdong and Liu, Shujie and Zhang, Tianhua and Chen, Youjun and Li, Jinyu and Meng, Helen M.},
  title = {{EmotionThinker}: Prosody-Aware Reinforcement Learning for Explainable Speech Emotion Reasoning},
  booktitle = {{ICLR}},
  year = {2026},
}

@inproceedings{xie2025audioreasoner,
  author = {Xie, Zhifei and Lin, Mingbao and Liu, Zihang and Wu, Pengcheng and Yan, Shuicheng and Miao, Chunyan},
  title = {{Audio-Reasoner}: Improving Reasoning Capability in Large Audio Language Models},
  booktitle = {{EMNLP}},
  pages = {23829--23851},
  year = {2025},
}

@article{kim2021emoberta,
  author = {Kim, Taewoon and Vossen, Piek},
  title = {{EmoBERTa}: Speaker-Aware Emotion Recognition in Conversation with {RoBERTa}},
  journal = {arXiv preprint arXiv:2108.12009},
  year = {2021},
}

@inproceedings{wang2020structbert,
  author = {Wang, Wei and Bi, Bin and Yan, Ming and Wu, Chen and Xia, Jiangnan and Bao, Zuyi and Peng, Liwei and Si, Luo},
  title = {{StructBERT}: Incorporating Language Structures into Pre-Training for Deep Language Understanding},
  booktitle = {{ICLR}},
  year = {2020},
}

@article{qwen2024qwen25,
  author = {{Qwen Team}},
  title = {{Qwen2.5} Technical Report},
  journal = {arXiv preprint arXiv:2412.15115},
  year = {2024},
}

@article{qwen2025qwen3,
  author = {{Qwen Team}},
  title = {{Qwen3} Technical Report},
  journal = {arXiv preprint arXiv:2505.09388},
  year = {2025},
}

@article{deepseekai2025deepseekv32,
  author = {{DeepSeek-AI}},
  title = {{DeepSeek-V3.2}: Pushing the Frontier of Open Large Language Models},
  journal = {arXiv preprint arXiv:2512.02556},
  year = {2025},
}

@inproceedings{tan2024edtalk,
  author = {Tan, Shuai and Ji, Bin and Bi, Mengxiao and Pan, Ye},
  title = {{EDTalk}: Efficient Disentanglement for Emotional Talking Head Synthesis},
  booktitle = {{ECCV}},
  pages = {398--416},
  year = {2024}
}

@inproceedings{meng2026echomimicv3,
  author = {Meng, Rang and Wang, Yan and Wu, Weipeng and Zheng, Ruobing and Li, Yuming and Ma, Chenguang},
  title = {{EchoMimicV3}: 1.3B Parameters Are All You Need for Unified Multi-Modal and Multi-Task Human Animation},
  booktitle = {{AAAI}},
  volume = {40},
  number = {10},
  pages = {8008--8015},
  year = {2026},
}

@inproceedings{liu2024anitalker,
  author = {Liu, Tao and Chen, Feilong and Fan, Shuai and Du, Chenpeng and Chen, Qi and Chen, Xie and Yu, Kai},
  title = {{AniTalker}: Animate Vivid and Diverse Talking Faces through Identity-Decoupled Facial Motion Encoding},
  booktitle = {{ACM MM}},
  pages = {6696--6705},
  year = {2024},
}

@inproceedings{cui2024hallo2,
  author = {Cui, Jiahao and Li, Hui and Yao, Yao and Zhu, Hao and Shang, Hanlin and Cheng, Kaihui and Zhou, Hang and Zhu, Siyu and Wang, Jingdong},
  title = {{Hallo2}: Long-Duration and High-Resolution Audio-Driven Portrait Image Animation},
  booktitle = {{ICLR}},
  pages = {91659--91671},
  year = {2025},
}

@article{wang2024vexpress,
  author = {Wang, Cong and Tian, Kuan and Zhang, Jun and Guan, Yonghang and Luo, Feng and Shen, Fei and Jiang, Zhiwei and Gu, Qing and Han, Xiao and Yang, Wei},
  title = {{V-Express}: Conditional Dropout for Progressive Training of Portrait Video Generation},
  journal = {arXiv preprint arXiv:2406.02511},
  year = {2024},
}

@inproceedings{li2025ditto,
  author = {Li, Tianqi and Zheng, Ruobing and Yang, Minghui and Chen, Jingdong and Yang, Ming},
  title = {Ditto: Motion-Space Diffusion for Controllable Realtime Talking Head Synthesis},
  booktitle = {{ACM MM}},
  year = {2025},
}

@inproceedings{ji2025sonic,
  author = {Ji, Xiaozhong and Hu, Xiaobin and Xu, Zhihong and Zhu, Junwei and Lin, Chuming and He, Qingdong and Zhang, Jiangning and Luo, Donghao and Chen, Yi and Lin, Qin and Lu, Qinglin and Wang, Chengjie},
  title = {Sonic: Shifting Focus to Global Audio Perception in Portrait Animation},
  booktitle = {{CVPR}},
  pages = {193--203},
  year = {2025}
}

@article{li2024latentsync,
  author = {Li, Chunyu and Zhang, Chao and Xu, Weikai and Lin, Jingyu and Xie, Jinghui and Feng, Weiguo and Peng, Bingyue and Chen, Cunjian and Xing, Weiwei},
  title = {{LatentSync}: Taming Audio-Conditioned Latent Diffusion Models for Lip Sync with {SyncNet} Supervision},
  journal = {arXiv preprint arXiv:2412.09262},
  year = {2024},
}

@article{tu2025stableavatar,
  author = {Tu, Shuyuan and Pan, Yueming and Huang, Yinming and Han, Xintong and Xing, Zhen and Dai, Qi and Luo, Chong and Wu, Zuxuan and Jiang, Yu-Gang},
  title = {{StableAvatar}: Infinite-Length Audio-Driven Avatar Video Generation},
  journal = {arXiv preprint arXiv:2508.08248},
  year = {2025},
}

@article{wei2024aniportrait,
  author = {Wei, Huawei and Yang, Zejun and Wang, Zhisheng},
  title = {{AniPortrait}: Audio-Driven Synthesis of Photorealistic Portrait Animations},
  journal = {arXiv preprint arXiv:2403.17694},
  year = {2024},
}

@inproceedings{xu2025hunyuanportrait,
  author = {Xu, Zunnan and Yu, Zhentao and Zhou, Zixiang and Zhou, Jun and Jin, Xiaoyu and Hong, Fa-Ting and Ji, Xiaozhong and Zhu, Junwei and Cai, Chengfei and Tang, Shiyu and Lin, Qin and Li, Xiu and Lu, Qinglin},
  title = {{HunyuanPortrait}: Implicit Condition Control for Enhanced Portrait Animation},
  booktitle = {{CVPR}},
  pages = {15909--15919},
  year = {2025}
}

@inproceedings{li2026personalive,
  author = {Li, Zhiyuan and Pun, Chi-Man and Fang, Chen and Wang, Jue and Cun, Xiaodong},
  title = {{PersonaLive!}: Expressive Portrait Image Animation for Live Streaming},
  booktitle = {{CVPR}},
  pages = {18118--18128},
  year = {2026}
}

@inproceedings{zhao2025xnemo,
  author = {Zhao, Xiaochen and Xu, Hongyi and Song, Guoxian and Xie, You and Zhang, Chenxu and Li, Xiu and Luo, Linjie and Suo, Jinli and Liu, Yebin},
  title = {{X-NeMo}: Expressive Neural Motion Reenactment via Disentangled Latent Attention},
  booktitle = {{ICLR}},
  pages = {77070--77088},
  year = {2025},
}

@inproceedings{tan2025animatex,
  author = {Tan, Shuai and Gong, Biao and Wang, Xiang and Zhang, Shiwei and Zheng, Dandan and Zheng, Ruobing and Zheng, Kecheng and Chen, Jingdong and Yang, Ming},
  title = {{Animate-X}: Universal Character Image Animation with Enhanced Motion Representation},
  booktitle = {{ICLR}},
  year = {2025},
}

@inproceedings{zhang2025mimicmotion,
  author = {Zhang, Yuang and Gu, Jiaxi and Wang, Li-Wen and Wang, Han and Cheng, Junqi and Zhu, Yuefeng and Zou, Fangyuan},
  title = {{MimicMotion}: High-Quality Human Motion Video Generation with Confidence-Aware Pose Guidance},
  booktitle = {{ICML}},
  volume = {267},
  pages = {74896--74910},
  year = {2025}
}

@inproceedings{tu2025stableanimator,
  author = {Tu, Shuyuan and Xing, Zhen and Han, Xintong and Cheng, Zhi-Qi and Dai, Qi and Luo, Chong and Wu, Zuxuan},
  title = {{StableAnimator}: High-Quality Identity-Preserving Human Image Animation},
  booktitle = {{CVPR}},
  pages = {21096--21106},
  year = {2025}
}

@misc{wanvideo2025wan22,
  author = {{Wan Team}},
  title = {{Wan2.2}: Wan Open and Advanced Large-Scale Video Generative Models},
  year = {2025},
}

@article{bao2024vidu,
  author = {Bao, Fan and Xiang, Chendong and Yue, Gang and He, Guande and Zhu, Hongzhou and Zheng, Kaiwen and Zhao, Min and Liu, Shilong and Wang, Yaole and Zhu, Jun},
  title = {Vidu: A Highly Consistent, Dynamic and Skilled Text-to-Video Generator with Diffusion Models},
  journal = {arXiv preprint arXiv:2405.04233},
  year = {2024},
}

@article{klingteam2026motioncontrol,
  author = {{Kling Team}},
  title = {{Kling-MotionControl} Technical Report},
  journal = {arXiv preprint arXiv:2603.03160},
  year = {2026},
}

@misc{rvcboss2024gptsovitsv2,
  author = {{RVC-Boss}},
  title = {{GPT-SoVITS} v2},
  year = {2024},
}

@inproceedings{zhou2026indextts2,
  author = {Zhou, Siyi and Zhou, Yiquan and He, Yi and Zhou, Xun and Wang, Jinchao and Deng, Wei and Shu, Jingchen},
  title = {{IndexTTS2}: A Breakthrough in Emotionally Expressive and Duration-Controlled Auto-Regressive Zero-Shot Text-to-Speech},
  booktitle = {{AAAI}},
  volume = {40},
  number = {41},
  pages = {35139--35148},
  year = {2026},
}

@article{du2025cosyvoice3,
  author = {Du, Zhihao and Gao, Changfeng and Wang, Yuxuan and Yu, Fan and Zhao, Tianyu and Wang, Hao and Lv, Xiang and Wang, Hui and Ni, Chongjia and Shi, Xian and An, Keyu and Yang, Guanrou and Li, Yabin and Chen, Yanni and Gao, Zhifu and Chen, Qian and Gu, Yue and Chen, Mengzhe and Chen, Yafeng and Zhang, Shiliang and Wang, Wen and Ye, Jieping},
  title = {{CosyVoice 3}: Towards In-the-Wild Speech Generation via Scaling-Up and Post-Training},
  journal = {arXiv preprint arXiv:2505.17589},
  year = {2025},
}

@misc{openaudio2025s1,
  author = {{Fish Audio}},
  title = {{OpenAudio S1}},
  year = {2025},
}

@misc{elevenlabs2023multilingualv2,
  author = {{ElevenLabs}},
  title = {Eleven Multilingual v2},
  year = {2023},
}

@misc{inworld2026tts15,
  author = {{Inworld AI}},
  title = {{Inworld TTS-1.5}: Upgrading the \#1 Ranked {TTS} Model with Production-Grade Latency, Expression, and Stability},
  year = {2026},
}

@inproceedings{sun2023sparsemat,
  author = {Sun, Yanan and Tang, Chi-Keung and Tai, Yu-Wing},
  title = {Ultrahigh Resolution Image/Video Matting with Spatio-Temporal Sparsity},
  booktitle = {{CVPR}},
  pages = {14112--14121},
  year = {2023}
}

@inproceedings{yang2026matanyone2,
  author = {Yang, Peiqing and Zhou, Shangchen and Hao, Kai and Tao, Qingyi},
  title = {{MatAnyone 2}: Scaling Video Matting via a Learned Quality Evaluator},
  booktitle = {{CVPR}},
  pages = {37476--37485},
  year = {2026}
}

@inproceedings{lim2026videomama,
  author = {Lim, Sangbeom and Oh, Seoung Wug and Huang, Jiahui and Yoon, Heeji and Kim, Seungryong and Lee, Joon-Young},
  title = {{VideoMaMa}: Mask-Guided Video Matting via Generative Prior},
  booktitle = {{CVPR}},
  year = {2026}
}

@article{qin2020u2net,
  author = {Qin, Xuebin and Zhang, Zichen and Huang, Chenyang and Dehghan, Masood and Zaiane, Osmar R. and Jagersand, Martin},
  title = {{U$^2$-Net}: Going Deeper with Nested U-Structure for Salient Object Detection},
  journal = {Pattern Recognit.},
  volume = {106},
  pages = {107404},
  year = {2020},
}

@inproceedings{kim2022inspyrenet,
  author = {Kim, Taehun and Kim, Kunhee and Lee, Joonyeong and Cha, Dongmin and Lee, Jiho and Kim, Daijin},
  title = {Revisiting Image Pyramid Structure for High Resolution Salient Object Detection},
  booktitle = {{ACCV}},
  pages = {108--124},
  year = {2022}
}

@inproceedings{carion2025sam3,
  author = {Carion, Nicolas and Gustafson, Laura and Hu, Yuan-Ting and Debnath, Shoubhik and Hu, Ronghang and Suris Coll-Vinent, Didac and Ryali, Chaitanya and Alwala, Kalyan Vasudev and Khedr, Haitham and Huang, Andrew and Lei, Jie and Ma, Tengyu and Guo, Baishan and Kalla, Arpit and Marks, Markus and Greer, Joseph and Wang, Meng and Sun, Peize and R{\"a}dle, Roman and Afouras, Triantafyllos and Mavroudi, Effrosyni and Xu, Katherine and Wu, Tsung-Han and Zhou, Yu and Momeni, Liliane and Hazra, Rishi and Ding, Shuangrui and Vaze, Sagar and Porcher, Francois and Li, Feng and Li, Siyuan and Kamath, Aishwarya and Cheng, Ho Kei and Dollar, Piotr and Ravi, Nikhila and Saenko, Kate and Zhang, Pengchuan and Feichtenhofer, Christoph},
  title = {{SAM 3}: Segment Anything with Concepts},
  booktitle = {{ICLR}},
  pages = {138846--138923},
  year = {2026},
}

@article{zheng2024birefnet,
  author = {Zheng, Peng and Gao, Dehong and Fan, Deng-Ping and Liu, Li and Laaksonen, Jorma and Ouyang, Wanli and Sebe, Nicu},
  title = {Bilateral Reference for High-Resolution Dichotomous Image Segmentation},
  journal = {CAAI AIR},
  volume = {3},
  pages = {1--12},
  year = {2024},
}

@misc{resembleai2026resemblyzer,
  author = {{Resemble AI}},
  title = {Resemblyzer: A Python Package to Analyze and Compare Voices with Deep Learning},
  year = {2026},
}

@article{johnson2016sparse,
  title   = {Sparse Coding for Alpha Matting},
  author  = {Johnson, Jubin and Shahrian Varnousfaderani, Ehsan
             and Cholakkal, Hisham and Rajan, Deepu},
  journal = {IEEE TIP},
  volume  = {25},
  number  = {7},
  pages   = {3032--3043},
  year    = {2016},
}

@inproceedings{perazzi2016benchmark,
  title     = {A Benchmark Dataset and Evaluation Methodology for
               Video Object Segmentation},
  author    = {Perazzi, Federico and Pont-Tuset, Jordi and McWilliams, Brian
               and Van Gool, Luc and Gross, Markus
               and Sorkine-Hornung, Alexander},
  booktitle = {{CVPR}},
  pages     = {724--732},
  year      = {2016},
}

\end{document}